\documentclass[10pt,twocolumn,letterpaper]{article}

\usepackage{iccv}              

\usepackage{graphicx}
\usepackage{array} 
\usepackage{amsthm,amsmath,amssymb,mathtools}
\usepackage{mathrsfs}
\usepackage{multirow}
\usepackage{colortbl}
\usepackage{adjustbox}
\usepackage{makecell}
\usepackage[accsupp]{axessibility}  

\definecolor{iccvblue}{rgb}{0.21,0.49,0.74}
\usepackage[pagebackref,breaklinks,colorlinks,allcolors=iccvblue]{hyperref}

\def\paperID{9450} 
\def\confName{ICCV}
\def\confYear{2025}
\def\modelname{DISCO}

\title{Federated Continual Instruction Tuning}

\author{
  Haiyang Guo\textsuperscript{\rm 1,2}, Fanhu Zeng\textsuperscript{\rm 2,3}, Fei Zhu\textsuperscript{\rm 4}, Wenzhuo Liu\textsuperscript{\rm 2,3}, Da-Han Wang\textsuperscript{\rm 5}, \\ Jian Xu\textsuperscript{\rm 2,3}, Xu-Yao Zhang\textsuperscript{\rm 1,2,3}\thanks{Corresponding Author.}, Cheng-Lin Liu\textsuperscript{\rm 1,2,3} \\
  \textsuperscript{\rm 1}School of Advanced Interdisciplinary Sciences, UCAS \textsuperscript{\rm 2}MAIS, CASIA \\ \textsuperscript{\rm 3}School of Artificial Intelligence, UCAS \textsuperscript{\rm 4}Centre for Artificial Intelligence and Robotics, HKISI-CAS \\ \textsuperscript{\rm 5}FKLPRIU, School of Computer and Information Engineering, Xiamen University of Technology \\
  {\normalsize \{guohaiyang2023, zengfanhu2022, jian.xu\}@ia.ac.cn, zhfei2018@gmail.com, \{xyz, liucl\}@nlpr.ia.ac.cn}
}

\begin{document}
\maketitle

\begin{abstract}
A vast amount of instruction tuning data is crucial for the impressive performance of Large Multimodal Models (LMMs), but the associated computational costs and data collection demands during supervised fine-tuning make it impractical for most researchers. Federated learning~(FL) has the potential to leverage all distributed data and training resources to reduce the overhead of joint training. However, most existing methods assume a fixed number of tasks, while in real-world scenarios, clients continuously encounter new knowledge and often struggle to retain old tasks due to memory constraints. In this work, we introduce the Federated Continual Instruction Tuning~(FCIT) benchmark to model this real-world challenge. Our benchmark includes two realistic scenarios, encompassing four different settings and twelve carefully curated instruction tuning datasets. To address the challenges posed by FCIT, we propose a dynamic knowledge organization to effectively integrate updates from different tasks during training and subspace selective activation to allocate task-specific output during inference. Extensive experimental results demonstrate that our proposed method significantly enhances model performance across varying levels of data heterogeneity and catastrophic forgetting. Code and dataset are released at \url{https://github.com/Ghy0501/FCIT}.
\end{abstract}

\vspace{-10pt}
\section{Introduction}

Large Multimodal Models~(LMMs)~\cite{bai2023qwen, liu2024visual, lu2024deepseek}, which integrate Large Language Model~\cite{brown2020language, jiang2023mistral, touvron2023llama, sun2024moss} with a visual encoder and multimodal projector to bridge visual and textual modalities, have exhibited impressive visual understanding and complex reasoning abilities. A crucial factor in this success is the supervised fine-tuning of LMMs using huge and diverse visual instruction-following data~\cite{liu2024visual} to align with human preferences. However, collecting such vast amounts of training data and computational resources for joint fine-tuning is impractical for most researchers. Federated Learning~(FL)~\cite{mcmahan2017communication, li2020federated, sani2024future, lin2023federated}, as a decentralized paradigm, offers a viable alternative by leveraging distributed data and computational resources for local training while integrating local weights to produce a unified model. This paradigm accommodates constraints on storage and computation while ensuring privacy protection. 

\begin{figure*}[t]
    \centering
    \includegraphics[width=0.98\linewidth]{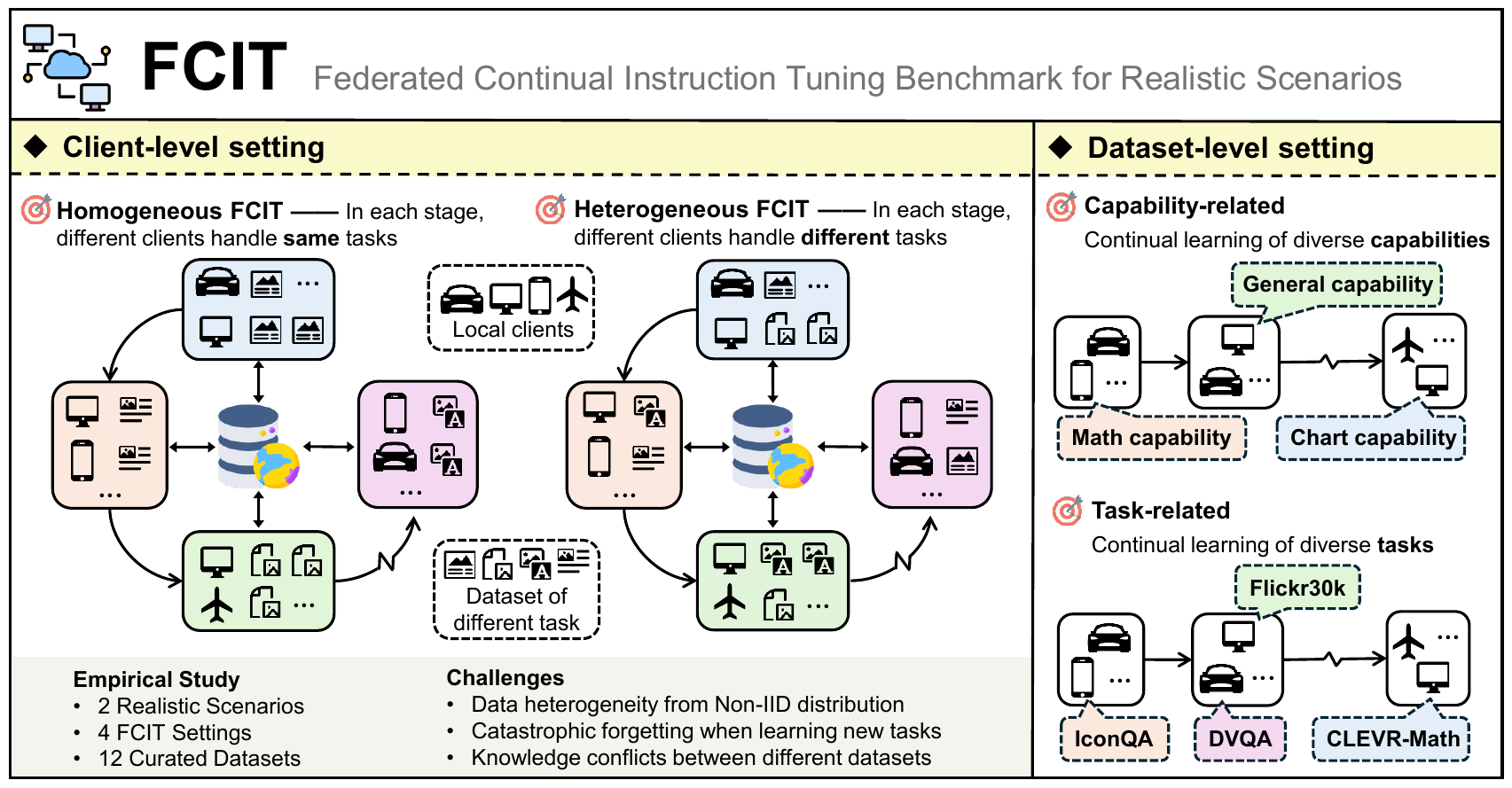}
    \vspace{-5pt}
    \caption{Overview of FCIT benchmark. FCIT encompasses 2 real-world scenarios, 4 FCIT settings, and 12 curated datasets, providing a comprehensive simulation of LMMs instruction-following training in real-world applications.}
    \label{fig:setting}
\vspace{-10pt}
\end{figure*}

Generally, most existing FL frameworks~\cite{ye2024openfedllm, zhang2024towards, chen2024feddat} are modeled in static, closed-world scenarios~\cite{zhang2020towards, bendale2015towards}, where a fixed set of tasks is predefined and unchanged. However, real-world applications are dynamic~\cite{zhu2024open, zhang2020towards}, requiring models to continuously acquire new knowledge while retaining previously learned tasks. Taking a realistic health emergency as an example, different hospitals act as local clients when a major disease outbreak occurs. Large hospitals can leverage FL to collaboratively train on their case data, building a comprehensive virus knowledge base, while smaller clinics update their own cases using this global knowledge. Over time, both large and small hospitals need to keep updating the global knowledge base through continual learning methods to cope with the event. Additionally, the emergence of new strains necessitates the simultaneous integration of knowledge and response strategies to minimize potential losses. In this scenario, traditional FL methods struggle with newly arrived knowledge, while Continual Learning (CL)~\cite{guo2025comprehensive, guo2025hide, zeng2024modalprompt, zhao2025mllm, liu2025llava} methods alone do not facilitate knowledge sharing between clients. Only an organic integration of both can address this real-world problem. Recently, numerous Federated Continual Learning methods have emerged to address this challenge in traditional image classification tasks~\cite{dong2022federated, guo2024pilora, zhang2023target}. However, their methodological and task-setting limitations make them insufficient for current LMM applications. Therefore, a more comprehensive benchmark is needed to better simulate the practical application of LMMs in real-world scenarios.

In this work, we first establish a \emph{Federated Continual Instruction Tuning}~(\emph{FCIT}) benchmark to fill this gap. Specifically, we simulate two client-level realistic scenarios: \textbf{(i) Homogeneous FCIT} refers to an FL system learning a series of instruction tuning data sequentially, with different clients learning the same task at each stage. \textbf{(ii) Heterogeneous FCIT}, on the other hand, requires the FL system to collaborate on different tasks simultaneously, with different clients potentially learning different tasks in the same stage. Building on this, we define two settings of datasets for each scenario: \textbf{Capability-related} and \textbf{Task-related}. The former evaluates the model's ability to integrate multiple dimensions of instruction following in a short period, while the latter assesses the model's performance during long phases of continual learning. For dataset selection, we curate twelve instruction tuning datasets unseen during LMM fine-tuning or with low zero-shot performance, preventing information leakage~\cite{kim2023learnability, guo2024desire}. In addition, we introduce varying degrees of data heterogeneity for each setting to challenge the model's performance in the non-IID situation~\cite{li2022federated}. An illustration of our FCIT benchmark is provided in~\Cref{fig:setting}. To the best of our knowledge, this is the first work to introduce a comprehensive benchmark for federated learning of LMM in a continual learning setting.

To effectively address the challenges posed by FCIT, we propose a novel \emph{\underline{D}ynamic knowledge organ\underline{I}zation} and \emph{\underline{S}ubspace sele\underline{C}tive activati\underline{O}n}~(\emph{\modelname{}}) framework. Specifically, we identify FCIT challenges into two types: conflicts between different tasks within the same stage and conflicts between old and new tasks across different stages. The former can cause catastrophic forgetting by altering the parameter space of previous tasks when learning new ones without access to past data, while the latter requires the model to integrate and organize knowledge from different tasks to derive unified representations. Therefore, we first propose \emph{Dynamic Knowledge Organization}~(DKO), which leverages a dynamic cache at the global server to store task-specific parameters. Using a unique identity token matching mechanism, it systematically organizes knowledge for different tasks into corresponding subspaces within the cache, effectively mitigating two types of conflicts. To better utilize the organized task subspaces in the dynamic cache, we introduce \emph{Subspace Selective Activation}~(SSA), which selectively activates subspaces relevant to the test input while filtering out irrelevant outputs, leading to significant performance improvements. Consequently, these designs enable our framework to efficiently tackle the data heterogeneity and catastrophic forgetting in FCIT. In summary, our major contributions are:
\begin{itemize}
    \item We present the first Federated Continual Instruction Tuning~(FCIT) benchmark designed for real-world scenarios, providing a comprehensive evaluation of LMMs to continuously learn new knowledge using distributed data and training resources in real applications.
    \item We propose a novel \modelname{} framework that integrates dynamic knowledge organization and subspace selective activation to efficiently address data heterogeneity and catastrophic forgetting in FCIT settings.
    \item Extensive experiments demonstrate that our method significantly enhances model performance under data heterogeneity while minimizing the catastrophic forgetting, and achieves state-of-the-art performance.
\end{itemize}

\section{Related Work}
\textbf{Large Multimodal Models.} With the grand unification of Large Language Models~(LLMs)~\cite{brown2020language, touvron2023llama, jiang2023mistral} for various NLP tasks, Large Multimodal Models~(LMMs)~\cite{zhu2023minigpt, dai2023instructblip, bai2023qwen, liu2024visual, lu2024deepseek} emerge by extending LLMs to combine with visual encoder and multimodal projectors, demonstrating exceptional visual understanding and complex reasoning abilities. To better align with human preferences, these LMMs typically undergo further fine-tuning on extensive instruction-following data, ensuring they meet the demands of real-world applications~\cite{liu2024visual}. However, in real-world scenarios, improving the performance of LMMs on new downstream tasks becomes a significant challenge without access to sufficient training data and computational resources. In this paper, we introduce the first Federated Continual Instruction Tuning~(FCIT) benchmark to bridge this gap from the perspective of distributed training and continual learning.

\noindent
\textbf{Federated Continual Learning.} In classical vision tasks, Federated Continual Learning~(FCL)~\cite{yang2024federated} aims to adapt the global model to new data while maintaining the knowledge of the old task. From the perspective of LMMs, research in this setting faces two main challenges: (1) Most of the methods are designed for traditional vision tasks~(\emph{i.e.,} image classification). For instance, MFCL~\cite{babakniya2024data} employs a generative model to synthesize images of previously learned classes, thereby mitigating forgetting. PILoRA~\cite{guo2024pilora} introduces a prototype re-weight module to address the classifier bias caused by data heterogeneity and obtain unified knowledge through LoRA~\cite{hu2021lora} fusion. Despite progress in classification tasks, complex designs remain challenging to adapt for LMMs research. (2) FCL follows a single data composition by splitting a dataset~(\eg ImageNet~\cite{deng2009imagenet}) into different tasks based on classes, whereas LMMs face greater challenges in continual learning due to their diverse tasks with varying styles~\cite{chen2024coin, zeng2024modalprompt, cao2024continual}.

This work pioneers a Federated Continual Instruction Tuning setup for LMMs and establishes diverse scenarios to simulate real-world applications comprehensively. Notably, AFCL~\cite{shenaj2023asynchronous} is most relevant to our work, as it enables clients to continuously learn multiple tasks in different orders and asynchronous time slots. However, it is specifically designed for image classification tasks, whereas our study focuses on the more widely used LMM.

\section{Problem Formulation}
\subsection{Preliminaries}

\textbf{Instruction tuning} enhances LMM's ability to understand and execute human instructions by performing supervised fine-tuning of pre-trained models on extensive datasets comprising instructions and responses. Formally, the instruction data $\mathcal{D} = \{ (\textbf{x}_{v}^{j}, \textbf{x}_{ins}^{j}, \textbf{x}_{res}^{j})_{j=1}^{N}\}$ consists of image input $\textbf{x}_{v}$, instruction $\textbf{x}_{ins}$ and response $\textbf{x}_{res}$, where $N$ represents the total number of samples. For clarity, given a simple image-instruction pair with a response of length $L$, the objective of an LMM is to predict the next token autoregressively based on all preceding tokens:
\begin{equation}
        p(\textbf{x}_{res}|\textbf{x}_v, \textbf{x}_{ins}) = \prod_{i=1}^{L}p_{\theta}(x_i|\textbf{x}_v, \textbf{x}_{ins},\textbf{x}_{res, <i}),
\label{eq:lmm}
\end{equation}
where $\theta$ denotes the trainable parameters during fine-tuning, $\textbf{x}_{\text{res},<i}$ denotes all response tokens preceding the current prediction token $x_i$. Then, the loss function of fine-tuning LMMs can be expressed as:
\begin{equation}
    \mathcal{L}_{\theta} = -\frac{1}{N}\sum_{j=1}^{N}\sum_{i=1}^{L_j}\log p_{\theta}(x_{i}^{j}|\textbf{x}_{v}^{j},\textbf{x}_{ins}^{j},\textbf{x}_{res,<i}^{j}).
\end{equation}

\noindent
\textbf{Federated learning} framework typically comprises a global server and several local clients, all employing the same LMM with a shared homogeneous model architecture. In each communication round\footnote{For clarification, we standardize the communication round in~\Cref{sec:DISCO} to 1.}, local clients train their models on own data and upload the updated weights to the global server for aggregation, enabling collaborative optimization of the global model while preserving data privacy.

Considering the excessive communication overhead of transferring the entire LMM between clients and the global server, we adopt LoRA~\cite{hu2021lora} for efficient fine-tuning, balancing training overhead and implementation cost~\cite{ye2024openfedllm, zhang2024towards}. Specifically, for a weight matrix $\textbf{W}_0 \in \mathbb{R}^{d\times k}$, LoRA decomposes parameter updates $\Delta \textbf{W}$ during fine-tuning into two low-rank subspaces:
\begin{equation}
    \textbf{W} = \textbf{W}_0 + \Delta \textbf{W} = \textbf{W}_0 + \textbf{B} \textbf{A},
\end{equation}
where $\textbf{B}\in \mathbb{R}^{d\times r}$, $\textbf{A}\in \mathbb{R}^{r\times k}$ and $r \ll \min{\{d,k\}}$.

\noindent
\textbf{Continual learning} aims to minimize the loss on the current task while retaining knowledge from previous tasks. Formally, given a sequence of datasets $\mathcal{D}_1, \mathcal{D}_2,\cdots, \mathcal{D}_T$, the optimization objective at task $t$ is:
\begin{equation}
\begin{aligned}
    \min_{\theta} \mathcal{L}(\theta) = \mathbb{E}_{(x,y) \sim \mathcal{D}_t}[\mathcal{L}(f_{\theta_{t}}({x}),y)] + \sum_{i=1}^{t-1}\epsilon_i, \\
    \text{s.t.}~ \mathbb{E}_{(x,y) \sim \mathcal{D}_i}[\mathcal{L}(f_{\theta_{t}}({x}),y) - \mathcal{L}(f_{\theta_{t-1}}({x}),y)] \leq \epsilon_i, \\
    \epsilon_i \geq 0; \forall i \in [1,\cdots,t-1],
\end{aligned}
\label{eq:cl}
\end{equation}
where $\epsilon_i$ is a slack variable that allows a small increase in the loss from the old datasets, providing tolerance for minor forgetting while focusing on learning the current task. In Eq.~\ref{eq:cl}, $x$ and $y$ can be viewed as the multimodal inputs $(\textbf{x}_v, \textbf{x}_{ins})$ and the response $\textbf{x}_{res}$, respectively.

\subsection{Federated Continual Instruction Tuning}
\label{sec:FCIT}
As shown in~\Cref{fig:setting}, we integrate federated and continual learning for LMMs within a unified framework and propose two client-level realistic scenarios: (1) \textbf{Homogeneous FCIT}~(Hom-FCIT). In this scenario, clients sequentially learn a series of tasks, allowing the global server to continuously update its knowledge. Each client learns the same task at a given stage and can only access data from that stage. (2) \textbf{Heterogeneous FCIT}~(Het-FCIT). In real-world applications, different clients may learn different tasks within the same stage, enabling the global server to respond more rapidly to diverse instructions. This requires the model not only to coordinate the knowledge of different tasks learned by clients in the current stage but also to mitigate forgetting during the learning process.
For each scenario, we define two dataset-level settings: capability-related and task-related, detailed as follows.

\begin{itemize}
    \item \textbf{Capability-related.} Following the dataset construction and division in LLaVA-OneVision~\cite{li2024llava}, we classify the 12 datasets into 4 capabilities: \emph{General}, \emph{Math}, \emph{Chart}, and \emph{Other}. The \emph{General} capability includes A-OKVQA~\cite{schwenk2022okvqa}, ImageNet-R~\cite{hendrycks2021many}, Grounding~\cite{mao2016generation}, and IconQA~\cite{lu2021iconqa}; \emph{Math} comprises CLEVR-Math~\cite{dahlgren2022clevr}, super-CLEVR~\cite{li2023super}, and TabMWP~\cite{lu2022dynamic}; \emph{Chart} involves ArxivQA~\cite{li2024multimodal}, FigureQA~\cite{kahou2017figureqa}, and DVQA~\cite{kafle2018dvqa}; and \emph{Other} encompasses OCR-VQA~\cite{mishra2019ocr} and Flickr30k~\cite{flickrentitiesijcv}. We treat these four capabilities as different stages of continual learning, with each capability comprising a mixture of datasets.
    \item \textbf{Task-related.} To evaluate the performance of different methods in a long-phase continual learning situation, we selected 8 datasets: ImageNet-R, ArxivQA, IconQA, CLEVR-Math, OCR-VQA, Flickr30k, FigureQA, and super-CLEVR as distinct stages of continual learning.
\end{itemize}

In total, we propose 4 FCIT settings to evaluate different methods. Compared to existing works~\cite{ye2024openfedllm, zhang2024towards, chen2024coin, zeng2024modalprompt, cao2024continual}, we are the first to explore the organic integration of FL and CL in the context of LMM. We provide further illustrations on the settings and datasets in Appendix~\ref{appendix:A}.

\section{The Proposed Framework: \emph{DISCO}}
\label{sec:DISCO}

\begin{figure*}[t]
    \centering
    \includegraphics[width=0.98\linewidth]{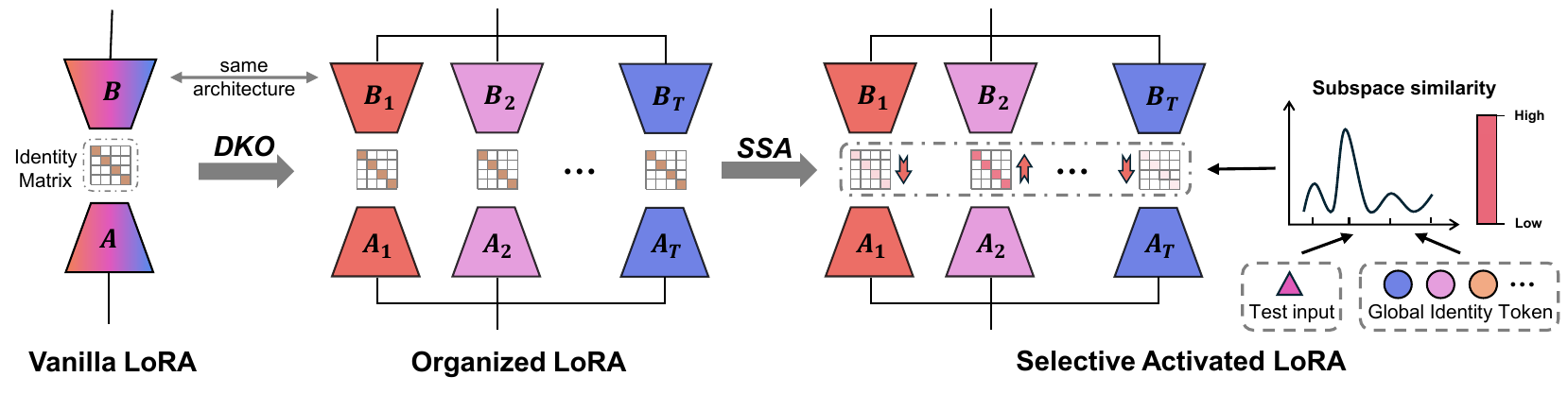}
    \vspace{-10pt}
    \caption{Illustration of the proposed SSA. Each subspace organized by DKO dynamically adjusts its output via the intrinsic activation matrix during inference, effectively filtering irrelevant information and enhancing model performance.}
    \label{fig:SSA}
\vspace{-15pt}
\end{figure*}

\textbf{Overview of the Method.} As shown in~\Cref{fig:DKO} and \ref{fig:SSA}, our method primarily consists of: \textbf{(1) Dynamic Knowledge Organization~(DKO)}, which dynamically integrates knowledge learned by different clients across stages during training, significantly reducing inter-task conflicts~(Section~\ref{sec:DKO}); and \textbf{(2) Subspace Selective Activation~(SSA)}, which selectively activates subspace outputs based on input features during inference, effectively filtering out irrelevant information~(Section~\ref{sec:SSA}). We name this framework \textbf{DISCO}.

\subsection{Dynamic Knowledge Organization}
\label{sec:DKO}
The core challenge of FCIT is enabling the global server to effectively harmonize knowledge learned by clients, addressing both the forgetting of previous knowledge when acquiring new tasks and the conflicts that arise from integrating knowledge from different tasks within the same stage. To this end, we propose maintaining a dynamic cache at the global server to organize task-specific knowledge uploaded by clients at different stages, thereby preventing both forgetting and conflicts between knowledge. Specifically, each task is assigned a dedicated parameter space to store and update its corresponding knowledge:
\begin{equation}
    \Delta \textbf{W} = \textbf{B} \textbf{A} \Leftrightarrow \underbrace{\{\textbf{B}_1 \textbf{A}_1 , \cdots , \textbf{B}_T \textbf{A}_T\}}_{\text{task-specific subspace}},
\label{eq:task-specific subspace}
\end{equation}
where $T$ is the number of tasks learned. At this point, it is crucial to aggregate the parameters uploaded by clients into their respective subspaces without privacy leakage.

\begin{figure}[t]
    \centering
    \includegraphics[width=0.98\linewidth]{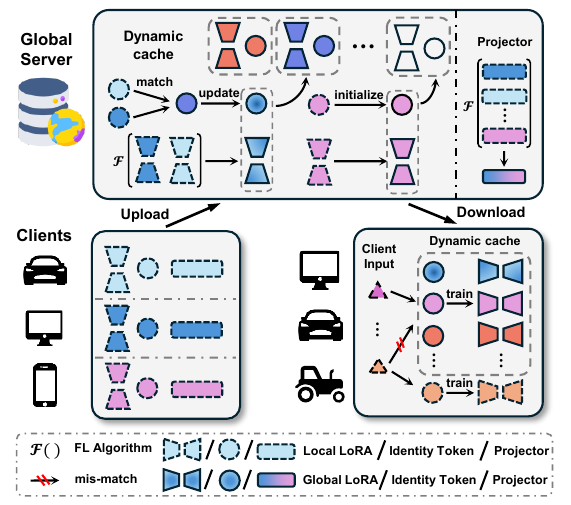}
    \vspace{-5pt}
    \caption{Illustration of the proposed DKO. Dynamic caches store the knowledge of each subspace while matching and updating through identity tokens.}
    \label{fig:DKO}
\vspace{-15pt}
\end{figure}

Inspired by the widespread use of prototypes in FL fields~\cite{tan2022fedproto, huang2023rethinking, tan2022federated}, we propose to distinguish knowledge across tasks by using the feature mean of each client's training data as an \textbf{Identity Token}. Specifically, considering the uniqueness of textual inputs in the visual instruction tuning task, we introduce a text encoder $f_{ins}$ for each client to extract the feature of the input instruction $\textbf{x}_{ins,j}^{t}$ during training, and take the mean value $\mathbf{\mu}_k^{t}$ as the local identity token of the $k$-th client on the task $t$ at the end of training:
\begin{equation}
    \mathbf{\mu}_k^{t} = \frac{1}{n_{k}^{t}}\sum_{j=1}^{n_{k}^{t}}f_{ins}(\textbf{x}_{ins,j}^{t}),
\end{equation}
where $n_{k}^{t}$ is the number of training samples of client $k$ at task $t$, and we use CLIP's text encoder~\cite{radford2021learning} in our experiments. Notably, the text encoder $f_{ins}$ remains frozen during training, which prevents additional training overhead.

Using the identity tokens, we first apply cosine similarity along with a threshold to match the uploaded local identity token $\mathbf{\mu}_j^{t}$ with the global identity token $\mathbf{\tilde{z}}_{i}$:
\begin{equation}
    \dfrac{\mathbf{\mu}_j^{t} \cdot \mathbf{\tilde{z}}_{i}}{\|\mathbf{\mu}_j^{t}\| \|\mathbf{\tilde{z}}_{i}\|} \geq \tau,
\label{eq:cos_sim}
\end{equation}
where $\tau$ denotes a pre-defined threshold. For paired local identity tokens, we update the corresponding global identity token. For mismatched local identity tokens, we pair them two by two using Eq.~(\ref{eq:cos_sim}) and initialize new global identity tokens. This process can be formalized as:
\begin{equation}
    \mathbf{\tilde{z}}_{i}^{*} =
    \begin{cases}
        \frac{n_{\mathbf{\tilde{z}}_{i}} \cdot \mathbf{\tilde{z}}_{i} + \sum_{j=1}^{m} n_j^t \cdot \mathbf{\mu}_j^t}{n_{\mathbf{\tilde{z}}_{i}} + \sum_{j=1}^{m} n_j^t}, & \text{if } \mathbf{\tilde{z}}_{i} \text{ exists,} \\
        \frac{\sum_{j=1}^{m} n_j^t \cdot \mathbf{\mu}_j^t}{\sum_{j=1}^{m} n_j^t}, & \text{if } \mathbf{\tilde{z}}_{i} \text{ doesn't exist,}
    \end{cases}
\end{equation}
where $m$, $n_{\mathbf{\tilde{z}}_{i}}$ and $n_j^t$ denote the number of matched local identity tokens, the number of samples used to form the previous $i$-th global identity token, and the number of samples from the local client $j$ at task $t$, respectively. Then, we leverage this matching process to guide the update or initialization of task-specific subspace $\{\mathbf{\theta}_i=\textbf{B}_i \textbf{A}_i\}$ in Eq.~(\ref{eq:task-specific subspace}):
\begin{equation}
    \theta_{i}^{*} = 
        \mathcal{F}(\theta_1^t,\cdots,\theta_m^t),
\end{equation}
where $\mathcal{F}$ denotes the FL algorithm~(\eg FedAvg~\cite{mcmahan2017communication}) used to aggregate local weights~\footnote{In \Cref{sec:further}, we implement more FL algorithms to demonstrate the compatibility of our method.}. As a result, we effectively prevent inter-task conflicts and integrate knowledge from different clients using the identity token matching mechanism.

After the global server completes the aggregation, it distributes the dynamic cache to each selected client. The client then matches the identity token~(Using Eq.~(\ref{eq:cos_sim})) of each subspace with its own training data, deciding whether to update the corresponding subspace or reinitialize a new one. The entire DKO process is illustrated in~\Cref{fig:DKO}.

\subsection{Subspace Selective Activation}
\label{sec:SSA}
Section~\ref{sec:DKO} effectively mitigates inter-task conflicts by disentangling and organizing complex knowledge into distinct subspaces. The key challenge then lies in how to leverage these task-specific subspaces during inference.

To address this, an intuitive method is to concatenate the subspaces of the dynamic cache in low-rank dimensions~\cite{wang2024flora, wang2023orthogonal}, enabling the integration of knowledge across all task spaces. However, this may introduce redundant information unrelated to the current task during inference, potentially compromising the model's output. For instance, when the desired answer is a simple word~(\eg, what is the object in the picture), knowledge from other subspaces, such as those used for generating long-form descriptions, can introduce unnecessary information, leading to responses that do not align with the given instruction. 

\begin{table*}[t]
    \centering
    \resizebox{0.95\linewidth}{!}{
    \begin{tabular}{>{\raggedright}p{2.5cm} *{12}{>{\centering\arraybackslash}p{1cm}}} 
        \toprule
        \multicolumn{1}{>{\raggedright}p{2.5cm}||}{Dataset setting} & \multicolumn{6}{c||}{Capability-related~(4 task)} & \multicolumn{6}{c}{Task-related~(8 task)} \\ 
        \midrule
        \multicolumn{1}{>{\raggedright}p{2.5cm}||}{Partition} & \multicolumn{2}{c|}{$\beta=0.5$} & \multicolumn{2}{c|}{$\beta=1.0$} & \multicolumn{2}{c||}{$\beta=5.0$} & \multicolumn{2}{c|}{$\beta=0.5$} & \multicolumn{2}{c|}{$\beta=1.0$} & \multicolumn{2}{c}{$\beta=5.0$} \\
        \midrule
        \multicolumn{1}{>{\raggedright}p{2.5cm}||}{Methods} & \multicolumn{1}{c}{Last} & \multicolumn{1}{c|}{Avg} & \multicolumn{1}{c}{Last} & \multicolumn{1}{c|}{Avg} & \multicolumn{1}{c}{Last} & \multicolumn{1}{c||}{Avg} & \multicolumn{1}{c}{Last} & \multicolumn{1}{c|}{Avg} & \multicolumn{1}{c}{Last} & \multicolumn{1}{c|}{Avg} & \multicolumn{1}{c}{Last} & \multicolumn{1}{c}{Avg} \\  
        \midrule 
        \multicolumn{1}{>{\raggedright}p{2.5cm}||}{Zero-shot} & \multicolumn{1}{c}{30.57} & \multicolumn{1}{c|}{-} & \multicolumn{1}{c}{30.57} & \multicolumn{1}{c|}{-} & \multicolumn{1}{c}{30.57} & \multicolumn{1}{c||}{-} & \multicolumn{1}{c}{29.08} & \multicolumn{1}{c|}{-} & \multicolumn{1}{c}{29.08} & \multicolumn{1}{c|}{-} & \multicolumn{1}{c}{29.08} & \multicolumn{1}{c}{-} \\  
        \multicolumn{1}{>{\raggedright}p{2.5cm}||}{Individual} & \multicolumn{1}{c}{61.75} & \multicolumn{1}{c|}{-} & \multicolumn{1}{c}{59.62} & \multicolumn{1}{c|}{-} & \multicolumn{1}{c}{60.27} & \multicolumn{1}{c||}{-} & \multicolumn{1}{c}{64.21} & \multicolumn{1}{c|}{-} & \multicolumn{1}{c}{63.85} & \multicolumn{1}{c|}{-} & \multicolumn{1}{c}{64.07} & \multicolumn{1}{c}{-} \\  
        \multicolumn{1}{>{\raggedright}p{2.5cm}||}{Centralized MTL} & \multicolumn{1}{c}{63.83} & \multicolumn{1}{c|}{-} & \multicolumn{1}{c}{63.83} & \multicolumn{1}{c|}{-} & \multicolumn{1}{c}{63.83} & \multicolumn{1}{c||}{-} & \multicolumn{1}{c}{66.60} & \multicolumn{1}{c|}{-} & \multicolumn{1}{c}{66.60} & \multicolumn{1}{c|}{-} & \multicolumn{1}{c}{66.60} & \multicolumn{1}{c}{-} \\  
        \midrule
        \multicolumn{1}{>{\raggedright}p{2.5cm}||}{Finetune} & \multicolumn{1}{c}{48.68} & \multicolumn{1}{c|}{60.05} & \multicolumn{1}{c}{49.67} & \multicolumn{1}{c|}{59.01} & \multicolumn{1}{c}{50.40} & \multicolumn{1}{c||}{58.08} & \multicolumn{1}{c}{46.24} & \multicolumn{1}{c|}{68.93} & \multicolumn{1}{c}{47.20} & \multicolumn{1}{c|}{68.79} & \multicolumn{1}{c}{48.00} & \multicolumn{1}{c}{69.97} \\  
        \multicolumn{1}{>{\raggedright}p{2.5cm}||}{EWC} & \multicolumn{1}{c}{49.12} & \multicolumn{1}{c|}{60.13} & \multicolumn{1}{c}{49.46} & \multicolumn{1}{c|}{59.28} & \multicolumn{1}{c}{49.89} & \multicolumn{1}{c||}{58.76} & \multicolumn{1}{c}{47.51} & \multicolumn{1}{c|}{69.04} & \multicolumn{1}{c}{47.92} & \multicolumn{1}{c|}{69.22} & \multicolumn{1}{c}{48.15} & \multicolumn{1}{c}{70.27} \\  
        \multicolumn{1}{>{\raggedright}p{2.5cm}||}{LwF} & \multicolumn{1}{c}{48.87} & \multicolumn{1}{c|}{60.32} & \multicolumn{1}{c}{50.02} & \multicolumn{1}{c|}{59.88} & \multicolumn{1}{c}{50.15} & \multicolumn{1}{c||}{59.06} & \multicolumn{1}{c}{47.89} & \multicolumn{1}{c|}{69.57} & \multicolumn{1}{c}{47.62} & \multicolumn{1}{c|}{69.14} & \multicolumn{1}{c}{48.26} & \multicolumn{1}{c}{70.30} \\ 
        \multicolumn{1}{>{\raggedright}p{2.5cm}||}{L2P} & \multicolumn{1}{c}{48.22} & \multicolumn{1}{c|}{59.79} & \multicolumn{1}{c}{49.56} & \multicolumn{1}{c|}{59.34} & \multicolumn{1}{c}{49.62} & \multicolumn{1}{c||}{59.25} & \multicolumn{1}{c}{47.30} & \multicolumn{1}{c|}{69.31} & \multicolumn{1}{c}{48.08} & \multicolumn{1}{c|}{69.65} & \multicolumn{1}{c}{48.42} & \multicolumn{1}{c}{70.16} \\ 
        \multicolumn{1}{>{\raggedright}p{2.5cm}||}{O-LoRA} & \multicolumn{1}{c}{51.65} & \multicolumn{1}{c|}{60.19} & \multicolumn{1}{c}{49.13} & \multicolumn{1}{c|}{57.93} & \multicolumn{1}{c}{50.29} & \multicolumn{1}{c||}{58.18} & \multicolumn{1}{c}{52.87} & \multicolumn{1}{c|}{71.54} & \multicolumn{1}{c}{49.87} & \multicolumn{1}{c|}{70.26} & \multicolumn{1}{c}{47.76} & \multicolumn{1}{c}{70.84} \\ 
        \multicolumn{1}{>{\raggedright}p{2.5cm}||}{M-LoRA} & \multicolumn{1}{c}{49.04} & \multicolumn{1}{c|}{60.08} & \multicolumn{1}{c}{50.39} & \multicolumn{1}{c|}{60.76} & \multicolumn{1}{c}{50.56} & \multicolumn{1}{c||}{58.06} & \multicolumn{1}{c}{50.68} & \multicolumn{1}{c|}{71.94} & \multicolumn{1}{c}{48.53} & \multicolumn{1}{c|}{71.58} & \multicolumn{1}{c}{48.38} & \multicolumn{1}{c}{71.21} \\ 
        \multicolumn{1}{>{\raggedright}p{2.5cm}||}{MoELoRA} & \multicolumn{1}{c}{49.69} & \multicolumn{1}{c|}{61.00} & \multicolumn{1}{c}{50.90} & \multicolumn{1}{c|}{60.18} & \multicolumn{1}{c}{50.43} & \multicolumn{1}{c||}{59.11} & \multicolumn{1}{c}{49.23} & \multicolumn{1}{c|}{70.96} & \multicolumn{1}{c}{49.02} & \multicolumn{1}{c|}{70.65} & \multicolumn{1}{c}{48.82} & \multicolumn{1}{c}{71.08} \\ \midrule
        \multicolumn{1}{>{\raggedright}p{2.5cm}||}{\textbf{DISCO}} & \multicolumn{1}{c}{\textbf{53.73}} & \multicolumn{1}{c|}{\textbf{62.00}} & \multicolumn{1}{c}{\textbf{55.47}} & \multicolumn{1}{c|}{\textbf{62.07}} & \multicolumn{1}{c}{\textbf{55.06}} & \multicolumn{1}{c||}{\textbf{60.53}} & \multicolumn{1}{c}{\textbf{57.69}} & \multicolumn{1}{c|}{\textbf{74.03}} & \multicolumn{1}{c}{\textbf{56.22}} & \multicolumn{1}{c|}{\textbf{73.03}} & \multicolumn{1}{c}{\textbf{55.58}} & \multicolumn{1}{c}{\textbf{72.64}} \\
        \bottomrule
    \end{tabular}}
    \caption{Last and Avg performance of different  methods on \textbf{Hom-FCIT} setting. The best performance is shown in bold.}
    \label{tab:Hom-FCIT}
\vspace{-10pt}
\end{table*}

\begin{figure*}[t]
    \centering
    \includegraphics[width=0.95\linewidth, height=0.25\linewidth]{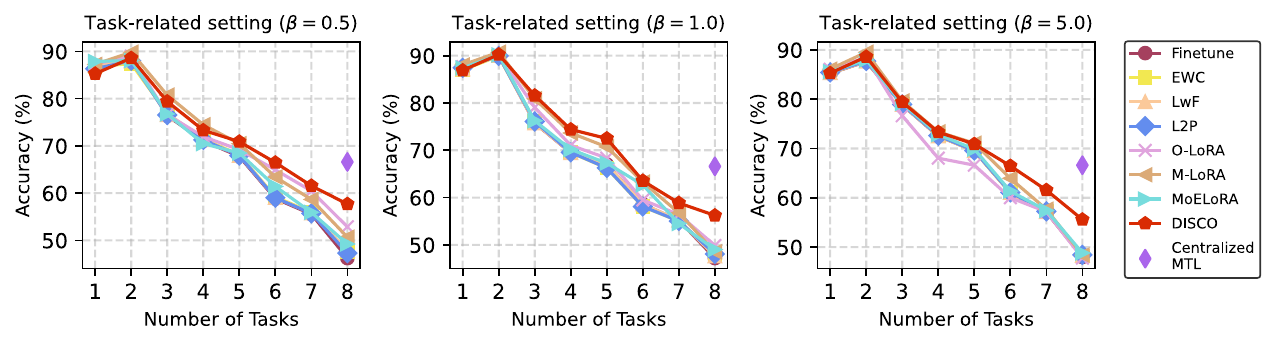}
    \vspace{-10pt}
    \caption{Performance curves of different methods in \textbf{Hom-FCIT} across seen tasks under varying data heterogeneity. We plot the average performance across all seen tasks at each stage.}
    \label{fig:Seq-FCIT}
\vspace{-10pt}
\end{figure*}

Drawing inspiration from LoRA's intrinsic space~\cite{wu2024mixture}, we propose subspace selective activation~(SSA) without additional training to filter out irrelevant subspace outputs, ensuring alignment between responses and instructions. In particular, A vanilla LoRA can be decomposed as the product of two low-rank subspaces~(\emph{i.e.,} $\textbf{A}$, $\textbf{B}$) and an intrinsic mixing matrix~$\mathcal{W}$:
\begin{equation}
    \Delta \textbf{W} = \textbf{B} \mathcal{W}  \textbf{A},
\end{equation}
where $\mathcal{W} \in \mathbb{R}^{r\times r}$ is typically an identity matrix and can thus be omitted. In this paper, we redefine $\mathcal{W}$ as the activation matrix that dynamically responds to the test input. Specifically, we treat $\mathcal{W}$ as the product of an identity matrix and an activation factor~(\emph{i.e.,} $\mathcal{W}=\alpha\cdot \mathbf{I}_{r\times r}$), where $\alpha=1$ denotes full activation, and $\alpha=0$ means that the output is fully masked. Therefore, for each subspace $\{\textbf{B}_1\textbf{A}_1,\cdots,\textbf{B}_T\textbf{A}_T\}$ in dynamic cache,  we can control the activation or inhibition of the corresponding output by adjusting its activation factor $\{\alpha_1,\cdots,\alpha_T\}$.

\begin{table*}[t]
    \centering
    \resizebox{0.95\linewidth}{!}{
    \begin{tabular}{>{\raggedright}p{2.5cm} *{12}{>{\centering\arraybackslash}p{1cm}}} 
        \toprule
        \multicolumn{1}{>{\raggedright}p{2.5cm}||}{Dataset setting} & \multicolumn{6}{c||}{Capability-related~(4 task)} & \multicolumn{6}{c}{Task-related~(8 task)} \\ 
        \midrule
        \multicolumn{1}{>{\raggedright}p{2.5cm}||}{Partition} & \multicolumn{2}{c|}{$\beta=0.5$} & \multicolumn{2}{c|}{$\beta=1.0$} & \multicolumn{2}{c||}{$\beta=5.0$} & \multicolumn{2}{c|}{$\beta=0.5$} & \multicolumn{2}{c|}{$\beta=1.0$} & \multicolumn{2}{c}{$\beta=5.0$} \\
        \midrule
        \multicolumn{1}{>{\raggedright}p{2.5cm}||}{Methods} & \multicolumn{1}{c}{Last} & \multicolumn{1}{c|}{Avg} & \multicolumn{1}{c}{Last} & \multicolumn{1}{c|}{Avg} & \multicolumn{1}{c}{Last} & \multicolumn{1}{c||}{Avg} & \multicolumn{1}{c}{Last} & \multicolumn{1}{c|}{Avg} & \multicolumn{1}{c}{Last} & \multicolumn{1}{c|}{Avg} & \multicolumn{1}{c}{Last} & \multicolumn{1}{c}{Avg} \\  
        \midrule 
        \multicolumn{1}{>{\raggedright}p{2.5cm}||}{Zero-shot} & \multicolumn{1}{c}{30.57} & \multicolumn{1}{c|}{-} & \multicolumn{1}{c}{30.57} & \multicolumn{1}{c|}{-} & \multicolumn{1}{c}{30.57} & \multicolumn{1}{c||}{-} & \multicolumn{1}{c}{29.08} & \multicolumn{1}{c|}{-} & \multicolumn{1}{c}{29.08} & \multicolumn{1}{c|}{-} & \multicolumn{1}{c}{29.08} & \multicolumn{1}{c}{-} \\  
        \multicolumn{1}{>{\raggedright}p{2.5cm}||}{Individual} & \multicolumn{1}{c}{61.75} & \multicolumn{1}{c|}{-} & \multicolumn{1}{c}{59.62} & \multicolumn{1}{c|}{-} & \multicolumn{1}{c}{60.27} & \multicolumn{1}{c||}{-} & \multicolumn{1}{c}{64.21} & \multicolumn{1}{c|}{-} & \multicolumn{1}{c}{63.85} & \multicolumn{1}{c|}{-} & \multicolumn{1}{c}{64.07} & \multicolumn{1}{c}{-} \\  
        \multicolumn{1}{>{\raggedright}p{2.5cm}||}{Centralized MTL} & \multicolumn{1}{c}{63.83} & \multicolumn{1}{c|}{-} & \multicolumn{1}{c}{63.83} & \multicolumn{1}{c|}{-} & \multicolumn{1}{c}{63.83} & \multicolumn{1}{c||}{-} & \multicolumn{1}{c}{66.60} & \multicolumn{1}{c|}{-} & \multicolumn{1}{c}{66.60} & \multicolumn{1}{c|}{-} & \multicolumn{1}{c}{66.60} & \multicolumn{1}{c}{-} \\  
        \midrule
        \multicolumn{1}{>{\raggedright}p{2.5cm}||}{Finetune} & \multicolumn{1}{c}{55.65} & \multicolumn{1}{c|}{55.82} & \multicolumn{1}{c}{56.34} & \multicolumn{1}{c|}{56.85} & \multicolumn{1}{c}{56.74} & \multicolumn{1}{c||}{57.15} & \multicolumn{1}{c}{58.04} & \multicolumn{1}{c|}{53.05} & \multicolumn{1}{c}{57.96} & \multicolumn{1}{c|}{54.22} & \multicolumn{1}{c}{58.17} & \multicolumn{1}{c}{54.87} \\  
        \multicolumn{1}{>{\raggedright}p{2.5cm}||}{EWC} & \multicolumn{1}{c}{55.21} & \multicolumn{1}{c|}{54.58} & \multicolumn{1}{c}{55.13} & \multicolumn{1}{c|}{55.76} & \multicolumn{1}{c}{55.69} & \multicolumn{1}{c||}{56.02} & \multicolumn{1}{c}{58.74} & \multicolumn{1}{c|}{53.76} & \multicolumn{1}{c}{57.96} & \multicolumn{1}{c|}{54.18} & \multicolumn{1}{c}{57.44} & \multicolumn{1}{c}{54.30} \\  
        \multicolumn{1}{>{\raggedright}p{2.5cm}||}{LwF} & \multicolumn{1}{c}{55.92} & \multicolumn{1}{c|}{56.18} & \multicolumn{1}{c}{56.42} & \multicolumn{1}{c|}{56.80} & \multicolumn{1}{c}{56.81} & \multicolumn{1}{c||}{57.26} & \multicolumn{1}{c}{58.82} & \multicolumn{1}{c|}{53.77} & \multicolumn{1}{c}{58.01} & \multicolumn{1}{c|}{54.34} & \multicolumn{1}{c}{58.22} & \multicolumn{1}{c}{54.78} \\ 
        \multicolumn{1}{>{\raggedright}p{2.5cm}||}{L2P} & \multicolumn{1}{c}{56.10} & \multicolumn{1}{c|}{56.63} & \multicolumn{1}{c}{56.76} & \multicolumn{1}{c|}{57.02} & \multicolumn{1}{c}{56.95} & \multicolumn{1}{c||}{57.36} & \multicolumn{1}{c}{58.84} & \multicolumn{1}{c|}{53.80} & \multicolumn{1}{c}{58.73} & \multicolumn{1}{c|}{54.67} & \multicolumn{1}{c}{58.39} & \multicolumn{1}{c}{54.66} \\ 
        \multicolumn{1}{>{\raggedright}p{2.5cm}||}{O-LoRA} & \multicolumn{1}{c}{58.24} & \multicolumn{1}{c|}{58.58} & \multicolumn{1}{c}{58.32} & \multicolumn{1}{c|}{58.65} & \multicolumn{1}{c}{58.60} & \multicolumn{1}{c||}{58.94} & \multicolumn{1}{c}{59.61} & \multicolumn{1}{c|}{54.55} & \multicolumn{1}{c}{59.74} & \multicolumn{1}{c|}{55.20} & \multicolumn{1}{c}{59.51} & \multicolumn{1}{c}{54.97} \\ 
        \multicolumn{1}{>{\raggedright}p{2.5cm}||}{M-LoRA} & \multicolumn{1}{c}{57.76} & \multicolumn{1}{c|}{58.02} & \multicolumn{1}{c}{57.65} & \multicolumn{1}{c|}{57.89} & \multicolumn{1}{c}{57.80} & \multicolumn{1}{c||}{58.17} & \multicolumn{1}{c}{59.76} & \multicolumn{1}{c|}{54.11} & \multicolumn{1}{c}{58.82} & \multicolumn{1}{c|}{54.03} & \multicolumn{1}{c}{59.35} & \multicolumn{1}{c}{54.89} \\ 
        \multicolumn{1}{>{\raggedright}p{2.5cm}||}{MoELoRA} & \multicolumn{1}{c}{57.68} & \multicolumn{1}{c|}{57.95} & \multicolumn{1}{c}{57.77} & \multicolumn{1}{c|}{58.00} & \multicolumn{1}{c}{58.15} & \multicolumn{1}{c||}{58.44} & \multicolumn{1}{c}{59.02} & \multicolumn{1}{c|}{54.25} & \multicolumn{1}{c}{59.14} & \multicolumn{1}{c|}{54.69} & \multicolumn{1}{c}{58.86} & \multicolumn{1}{c}{54.50} \\ \midrule
        \multicolumn{1}{>{\raggedright}p{2.5cm}||}{\textbf{DISCO}} & \multicolumn{1}{c}{\textbf{59.40}} & \multicolumn{1}{c|}{\textbf{60.01}} & \multicolumn{1}{c}{\textbf{59.94}} & \multicolumn{1}{c|}{\textbf{59.91}} & \multicolumn{1}{c}{\textbf{59.71}} & \multicolumn{1}{c||}{\textbf{60.16}} & \multicolumn{1}{c}{\textbf{62.08}} & \multicolumn{1}{c|}{\textbf{59.64}} & \multicolumn{1}{c}{\textbf{63.25}} & \multicolumn{1}{c|}{\textbf{61.99}} & \multicolumn{1}{c}{\textbf{62.78}} & \multicolumn{1}{c}{\textbf{60.87}} \\
        \bottomrule
    \end{tabular}}
    \caption{Last and Avg performance of different  methods on \textbf{Het-FCIT} setting. The best performance is shown in bold.}
    \label{tab:Het-FCIT}
\vspace{-10pt}
\end{table*}

\begin{figure*}[t]
    \centering
    \includegraphics[width=0.95\linewidth, height=0.25\linewidth]{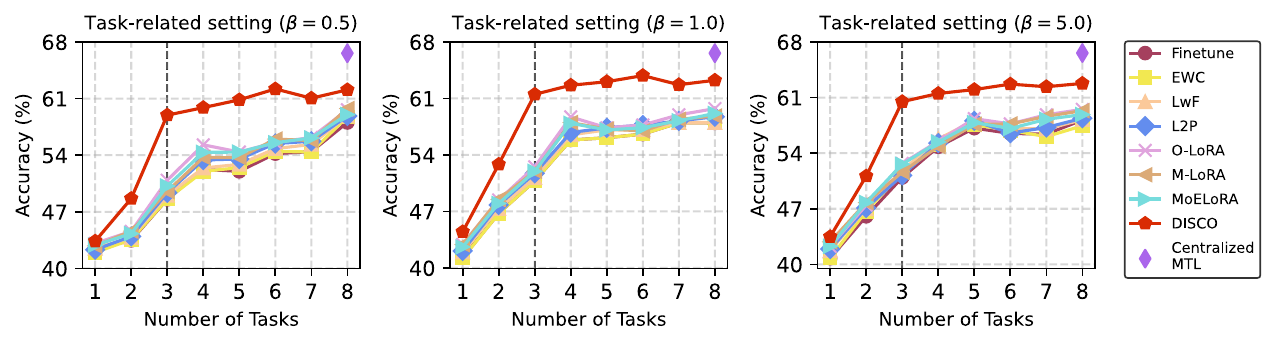}
    \vspace{-10pt}
    \caption{Performance curve of different methods in the \textbf{Het-FCIT} under varying degrees of data heterogeneity. We plot the average performance across all tasks at each stage. The black dashed line marks the stage where the model has encountered all tasks.}
    \label{fig:Dyn-FCIT}
\vspace{-10pt}
\end{figure*}

To provide better flexibility in assigning activation factors, we use global identity tokens and test input features extracted by the text encoder for similarity matching:
\begin{equation}
    s_i = \frac{\mathbf{\tilde{z}_i}\cdot f_{ins}(\textbf{x}_{ins}^{test})}{\|\mathbf{\tilde{z}_i}\|\cdot \|f_{ins}(\textbf{x}_{ins}^{test})\|},
\label{eq:cos-sim}
\end{equation}
where $s_i$ denotes the similarity between the $i$-th global identity token and the feature of the test instruction $\textbf{x}_{ins}^{test}$. The activation factor $\alpha_i$ is then computed by applying softmax normalization over all similarities:
\begin{equation}
    \alpha_i = \frac{\exp(s_i/\varepsilon)}{\sum_{j=1}^T \exp(s_j/\varepsilon)},
\label{eq:softmax}
\end{equation}
where $\varepsilon$ is the temperature coefficient. 

Overall, our proposed SSA can be formulated as follows:
\begin{equation}
    \begin{aligned}
        \Delta \textbf{W} &= \bar{\textbf{B}}\bar{\mathcal{W}}\bar{\textbf{A}} \\
        &= \bar{\textbf{B}}
        \begin{bmatrix}
        \alpha_1\cdot\mathbf{I}_{r\times r} & \mathbf{0}_{r\times r} & \cdots & \mathbf{0}_{r\times r} \\
        \mathbf{0}_{r\times r} & \alpha_2\cdot\mathbf{I}_{r\times r} & \cdots & \mathbf{0}_{r\times r} \\
        \vdots & \vdots & \ddots & \vdots \\
        \mathbf{0}_{r\times r} & \mathbf{0}_{r\times r} & \cdots & \alpha_T\cdot\mathbf{I}_{r\times r}
        \end{bmatrix}
        \bar{\textbf{A}},
    \end{aligned}
\end{equation}
where $\bar{\textbf{B}} \in \mathbb{R}^{d\times(r\cdot T)}$ represents the concatenation of all $\{\textbf{B}_1,\cdots,\textbf{B}_T\}$ in the dynamic cache along the low-rank dimension $r$, and $\bar{\textbf{A}} \in \mathbb{R}^{(r\cdot T) \times k}$ follows the same structure. The activation factors $\{\alpha_1,\cdots,\alpha_T\}$, computed via Eq.~(\ref{eq:cos-sim}) and Eq.~(\ref{eq:softmax}), control each subspace's output by amplifying matched subspaces while suppressing irrelevant ones. The schematic of SSA is provided in~\Cref{fig:SSA}.

\section{Experiments}
\subsection{Experimental Setup}

\textbf{Datasets.} The dataset composition is detailed in Section~\ref{sec:FCIT}. For the experimental setup, we implement two types of dataset-level settings: Capability-related and Task-related, based on the two client-level realistic scenarios, Hom-FCIT and Het-FCIT, respectively. At each stage, the own data of local clients is partitioned according to the Dirichlet distribution~\cite{li2022federated}, with three partitions $\beta$ for each setting to model different levels of data heterogeneity. More details are provided in Appendix~\ref{appendix:A}.

\noindent
\textbf{Baselines \& Evaluation Metrics.} We compare our method with continual learning methods such as LwF~\cite{li2017learning}, EWC~\cite{kirkpatrick2017overcoming}, L2P~\cite{wang2022learning}, O-LoRA~\cite{wang2023orthogonal}, MoELoRA~\cite{chen2024coin}, and also with the LoRA merging method from the federated continual learning approach PILoRA~\cite{guo2024pilora}, referred to as M-LoRA. We follow a rehearsal-free continual learning setting~\cite{zhu2021prototype}, where only the data for the current task is available. All comparison methods are carefully calibrated to ensure the fairness of evaluations.

For the evaluation metrics, we report the standard metrics to measure the model performance: \emph{\textbf{Last}} refers to the average result across all learned tasks after the completion of learning the final task. \emph{\textbf{Avg}} is based on Last, which tracks the performance of the learned tasks at each stage of the continual learning process and reports the average results.

\noindent
\textbf{Implementation Details.} We choose LLaVA-1.5-7b~\cite{liu2024visual} as the base LMM due to its structural simplicity and adopt its LoRA fine-tuning strategy, training only the LoRAs and multimodal projector. The LoRA rank $r$ at each stage is set to 8, with a learning rate of 2e-4, and it is embedded solely in the FFN layers of each block. The learning rate for the projector is set to 2e-5. For identity token extraction, we use the frozen CLIP text encoder~\cite{radford2021learning}. The threshold $\tau$ and temperature coefficient $\varepsilon$ are set to 0.9 and 0.05, respectively.

For the setting of FCIT, we set the epoch to 1 and communication rounds to 10 in each stage, and the Dirichlet distribution coefficients $\beta$ are set to $\{0.5, 1.0, 5.0\}$. In each round, the global server randomly selects 5 clients from a pool of 50 to participate in the training, and we use FedAvg~\cite{mcmahan2017communication} as the base FL aggregation algorithm.

\subsection{Main Results}

Results are shown in~\Cref{tab:Hom-FCIT},~\Cref{tab:Het-FCIT},~\Cref{fig:Seq-FCIT}, and \Cref{fig:Dyn-FCIT}. Under Hom-FCIT setting, our proposed DISCO achieves the best performance in both capability-related and task-related dataset settings, surpassing the best baseline by an average of \textbf{4.84\%} and \textbf{1.43\%} in the Last and Avg metrics, respectively. As illustrated in~\Cref{fig:Seq-FCIT}, DISCO effectively mitigates the forgetting of previous tasks while learning new ones, outperforming other methods in the challenging long-stage continual learning setting.

In Het-FCIT setting, we plot the upward curve of the average performance across all tasks in~\Cref{fig:Dyn-FCIT}. As shown in the figure. Before completing all tasks (Left of the black dotted line), our method maintains the fastest rate of improvement, significantly outperforming other methods. In the subsequent phases (Right of the black dashed line), it continues to consolidate previously learned knowledge and further enhance performance, demonstrating strong adaptability in dynamic real-world scenarios. On the Last and Avg metrics, our method outperforms the best comparison method by an average of \textbf{2.17\%} and \textbf{3.62\%}, respectively.

Additionally, our method demonstrates superior robustness to varying degrees of data heterogeneity, adapting well to diverse distributions and maintaining strong performance even as heterogeneity increases, highlighting its reliability in dynamic real-world scenarios.

\subsection{Ablation Study}

\begin{table}[t]
	\centering
        \resizebox{\linewidth}{!}{
	\begin{tabular}{c>{\raggedright}p{2.3cm} *{8}{>{\centering\arraybackslash}p{0.75cm}}}
		\toprule		
            & {\hspace{0.3em}} \multirow{2}{*}{Method} & \multicolumn{4}{c}{Hom-FCIT} & \multicolumn{4}{c}{Het-FCIT} \\ 
            \cline{3-10}
            \noalign{\smallskip} & & \multicolumn{1}{c}{Last} & $\Delta$ & \multicolumn{1}{c}{Avg} & $\Delta$ & \multicolumn{1}{c}{Last} & $\Delta$ & \multicolumn{1}{c}{Avg} & $\Delta$ \\
            \midrule
            \rowcolor{red!10}
            \multirow{3}{*}{(a)} & {\hspace{0.3em}} Text & \textbf{56.22} & 0.0 & \textbf{73.03} & 0.0 & \textbf{63.25} & 0.0 & \textbf{61.99} & 0.0\\
            & {\hspace{0.3em}} Image & 55.63 & \textcolor[rgb]{0, 0, 0.81}{-0.59} & 72.13 & \textcolor[rgb]{0, 0, 0.81}{-0.90} & 63.00 & \textcolor[rgb]{0, 0, 0.81}{-0.25} & 61.80 & \textcolor[rgb]{0, 0, 0.81}{-0.19} \\
            & {\hspace{0.3em}} Text \& Image & 55.96 & \textcolor[rgb]{0, 0, 0.81}{-0.26} & 72.68 & \textcolor[rgb]{0, 0, 0.81}{-0.35} & 63.02 & \textcolor[rgb]{0, 0, 0.81}{-0.23} & 61.78 & \textcolor[rgb]{0, 0, 0.81}{-0.21} \\
            \midrule
            \rowcolor{red!10}
            \multirow{4}{*}{(b)} & {\hspace{0.3em}} Softmax & \textbf{56.22} & 0.0 & \textbf{73.03} & 0.0 & \textbf{63.25} & 0.0 & \textbf{61.99} & 0.0\\
            & {\hspace{0.3em}} Concatenate & 51.74 & \textcolor[rgb]{0, 0, 0.81}{-4.48} & 69.20 & \textcolor[rgb]{0, 0, 0.81}{-3.83} & 60.36 & \textcolor[rgb]{0, 0, 0.81}{-2.89} & 59.88 & \textcolor[rgb]{0, 0, 0.81}{-2.11} \\
            & {\hspace{0.3em}} Cosine sim & 52.83 & \textcolor[rgb]{0, 0, 0.81}{-3.39} & 70.07 & \textcolor[rgb]{0, 0, 0.81}{-2.96} & 60.92 & \textcolor[rgb]{0, 0, 0.81}{-2.33} & 60.13 & \textcolor[rgb]{0, 0, 0.81}{-1.86} \\
             & {\hspace{0.3em}} Argmax & 55.74 & \textcolor[rgb]{0, 0, 0.81}{-0.48} & 72.07 & \textcolor[rgb]{0, 0, 0.81}{-0.96} & 62.88 & \textcolor[rgb]{0, 0, 0.81}{-0.37} & 61.42 & \textcolor[rgb]{0, 0, 0.81}{-0.57} \\
            \midrule
            \rowcolor{red!10}
            \multirow{3}{*}{(c)} & {\hspace{0.3em}} FFN & \textbf{56.22} & 0.0 & \textbf{73.03} & 0.0 & \textbf{63.25} & 0.0 & \textbf{61.99} & 0.0\\
            & {\hspace{0.3em}} Attn & 56.03 & \textcolor[rgb]{0, 0, 0.81}{-0.19} & 72.88 & \textcolor[rgb]{0, 0, 0.81}{-0.15} & 63.01 & \textcolor[rgb]{0, 0, 0.81}{-0.24} & 61.83 & \textcolor[rgb]{0, 0, 0.81}{-0.16} \\
            & {\hspace{0.3em}} FFN \& Attn & 56.30 & \textcolor[rgb]{0.81,0,0}{+0.08} & 72.96 & \textcolor[rgb]{0, 0, 0.81}{-0.07} & 63.18 & \textcolor[rgb]{0, 0, 0.81}{-0.07} & 62.04 & \textcolor[rgb]{0.81,0,0}{+0.05} \\
		\bottomrule
	\end{tabular}}
	\caption{Ablation studies on (a) identity token extraction         methods; (b) calculation of activation factors in SSA; (c) location of LoRA embedding. All experiments were conducted in the task-related setting with $\beta=1.0$.}
  \vspace{-15pt}
\label{tab:ablation}
\end{table}

The detailed ablation studies are provided in~\Cref{tab:ablation}. In this paper, we use CLIP’s text encoder to extract textual features as identity tokens. This section tests two alternatives: using only visual features from CLIP’s visual encoder, and combining visual and textual features. From~\Cref{tab:ablation}\textcolor{iccvblue}{a}, we observe that both alternatives degrade performance. This is potential because, in visual instruction tuning datasets, task similarities at the image level are higher than at the textual level (\eg CLEVR-Math and super-CLEVR), making visual features less effective than textual ones.

\Cref{tab:ablation}\textcolor{iccvblue}{b} shows the ablation studies on how to calculate the activation factors in SSA. Direct concatenation, where the activation factors of all subspaces are set to 1.0, leads to significant performance degradation due to unfiltered outputs. Using cosine similarity~(Eq.(\ref{eq:cos-sim})) to filter the outputs also fails to solve the issue. While selecting the largest similarity (Argmax) can filter out irrelevant information, it risks fully activating a subspace that, though similar, is not directly related to the current task. This can negatively impact the output, as there are inherent similarities between the textual information of different tasks. In contrast, our method normalizes the cosine similarity to better focus on the relevant task output, achieving optimal performance.

We also test the location of LoRA embedding. As can be seen in~\Cref{tab:ablation}\textcolor{iccvblue}{c}, embedding in the attention layer alone proved less effective than in the FFN layer. Moreover, embedding LoRA in every linear layer did not offer significant improvement and resulted in higher parameter transmission. Therefore, we choose to embed LoRA only in the FFN layer as a balanced solution. We provide more ablation and visualization results in Appendix~\ref{appendix:B}.

\subsection{Further Analysis}
\label{sec:further}

\textbf{Compatible with other FL algorithms.} In this paper, we use the classical FedAvg as the FL algorithm for global server aggregation of local weights. Additionally, we implement other federated learning algorithms, including FedAvgM~\cite{hsu2019measuring}, FedAdam~\cite{reddi2020adaptive}, FedAdagrad~\cite{reddi2020adaptive}, and FedYogi~\cite{reddi2020adaptive}, to extend our framework. As shown in~\Cref{tab:fl_methods}, FedAvg achieves the best average performance.

\begin{table}[t]
	\centering
        \resizebox{0.85\linewidth}{!}{
	\begin{tabular}{>{\raggedright}p{2.0cm} *{4}{>{\centering\arraybackslash}p{2.0cm}}}
		\toprule		
            \multirow{2}{*}{Method} & \multicolumn{2}{c}{Hom-FCIT} & \multicolumn{2}{c}{Het-FCIT} \\ 
            \cline{2-5}
            \noalign{\smallskip} & \multicolumn{1}{c}{Last} & \multicolumn{1}{c|}{Avg} & \multicolumn{1}{c}{Last} & \multicolumn{1}{c}{Avg} \\
            \midrule
            FedAvg & \multicolumn{1}{c}{\textbf{56.22}} & \multicolumn{1}{c|}{73.03} & \multicolumn{1}{c}{\textbf{63.25}} & \multicolumn{1}{c}{\textbf{61.99}} \\
            FedAvgM & \multicolumn{1}{c}{56.07} & \multicolumn{1}{c|}{72.59} & \multicolumn{1}{c}{62.47} & \multicolumn{1}{c}{61.10} \\
            FedAdam & \multicolumn{1}{c}{55.76} & \multicolumn{1}{c|}{72.91} & \multicolumn{1}{c}{62.78} & \multicolumn{1}{c}{61.35} \\
            FedAdagrad & \multicolumn{1}{c}{56.02} & \multicolumn{1}{c|}{72.80} & \multicolumn{1}{c}{62.55} & \multicolumn{1}{c}{61.41} \\
            FedYogi & \multicolumn{1}{c}{56.11} & \multicolumn{1}{c|}{\textbf{73.16}} & \multicolumn{1}{c}{62.88} & \multicolumn{1}{c}{61.67} \\
		\bottomrule
	\end{tabular}}
	\caption{Results of different FL algorithms. All experiments were conducted in the task-related setting with $\beta=1.0$.}
  \vspace{-10pt}
\label{tab:fl_methods}
\end{table}

\begin{figure}[t]
    \centering
    \includegraphics[width=0.8\linewidth]{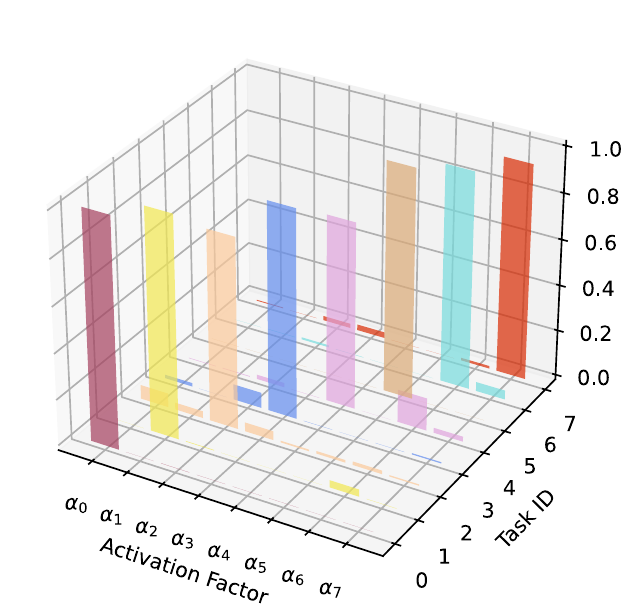}
    \vspace{-5pt}
    \caption{Visualization of activation factors during inference.}
    \label{fig:activation}
\vspace{-10pt}
\end{figure}

\noindent
\textbf{Visualization of activation factors.} In~\Cref{fig:activation}, we plot the responses of activations to the test inputs of different tasks. As observed, only the activator corresponding to the current task is responsive, effectively activating the relevant subspace, while the others remain largely inhibited. This confirms the effectiveness of our proposed SSA.

\section{Conclusion}
We have explored for the first time the integration of federated learning and continual learning for the instruction tuning of LMMs, addressing the real-world challenge of dynamically acquiring new knowledge through distributed training resources and data. Our proposed FCIT benchmark encompasses 2 real-world scenarios, 4 distinct settings, and 12 curated datasets, providing a comprehensive evaluation of different methods. Additionally, we introduce the DISCO framework, which leverages dynamic knowledge organization~(DKO) to decompose inter-task conflicts and subspace selective activation~(SSA) to assign task-relevant outputs while suppressing irrelevant information. Extensive experiments demonstrate that our approach significantly improves the model’s ability to learn new knowledge and handle data heterogeneity in real-world scenarios.

\section*{Acknowledgments}
This work was supported by the National Science and Technology Major Project (2022ZD0116500), National Natural Science Foundation of China (62222609, 62320106010), CAS Project for Young Scientists in Basic Research (YSBR-083), Strategic Priority Research Program of Chinese Academy of Sciences under Grant (XDA0480200), Major Science and Technology Plan Project on the Future Industry Fields of Xiamen City (3502Z20241027), Unveiling and Leading Projects of Xiamen (3502Z20241011) and the InnoHK program. 

{
    \small
    \bibliographystyle{ieeenat_fullname}
    \bibliography{main}
}

\clearpage
\setcounter{page}{1}
\setcounter{figure}{0}
\setcounter{table}{0}
\renewcommand{\thefigure}{A\arabic{figure}}
\renewcommand{\thetable}{A\arabic{table}}
\maketitlesupplementary
\appendix

\section{More Details of FCIT Benchmatk}
\label{appendix:A}

\begin{figure}[t]
    \centering
    \includegraphics[width=0.99\linewidth]{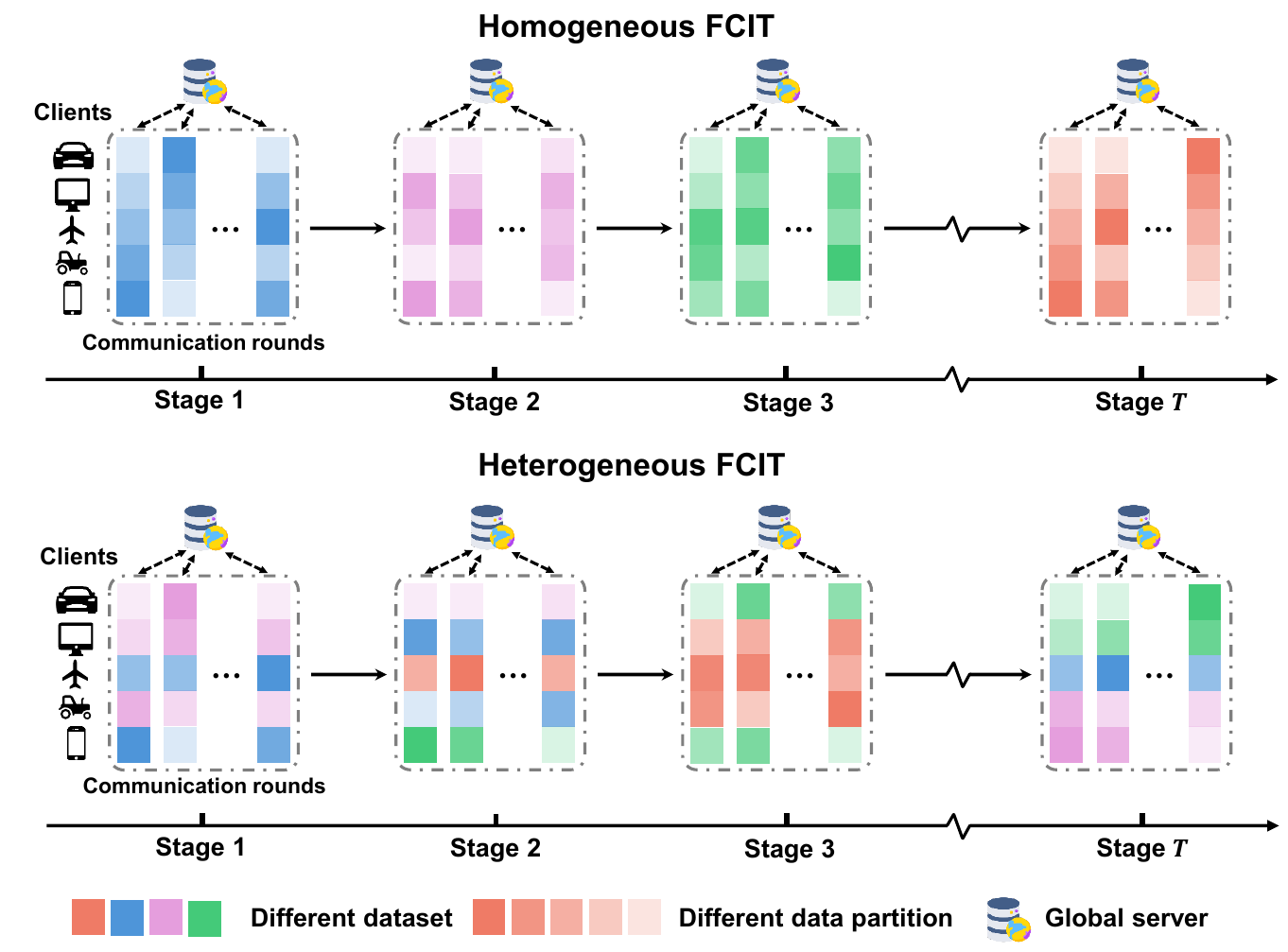}
    \vspace{-5pt}
    \caption{Illustration of Homogeneous FCIT and Heterogeneous FCIT settings.}
    \label{fig:flowchat}
\vspace{-10pt}
\end{figure}

\subsection{Illustration display of FCIT}
To better understand the proposed Homogeneous FCIT and Heterogeneous FCIT settings, we further provide an illustration detailing their entire process.

As shown in~\Cref{fig:flowchat}, we use different colors to distinguish datasets and assume the continual learning stage lasts for $T$ steps.  In \textbf{Homogeneous FCIT} setting, the own data of the selected local clients in each stage belong to the same dataset, which exists in different proportions in different clients. Each stage comprises multiple communication rounds (set to 10 in our experiments), during which clients train the model on their own data (with 1 epoch per round), upload weights to the global server for aggregation, and receive the aggregated weights for the next round. Homogeneous FCIT setting extends continual instruction tuning to a federated learning setting with a non-IID data distribution, posing greater challenges for traditional methods.

\textbf{Heterogeneous FCIT} setting extends the former by allowing each client's data to come from different datasets in each stage. This requires the global server to mitigate catastrophic forgetting across stages while resolving conflicts among datasets within the same stage. This setting is common in real-world scenarios, such as healthcare systems that need to simultaneously manage multiple disease outbreaks while continuously updating to track their progression and mitigate social risks, and our benchmark effectively addresses this real-world need, providing a comprehensive evaluation framework for such dynamic challenges.

\subsection{Visualization of data heterogeneous}
\begin{figure}[h]
    \centering
    \includegraphics[width=0.99\linewidth]{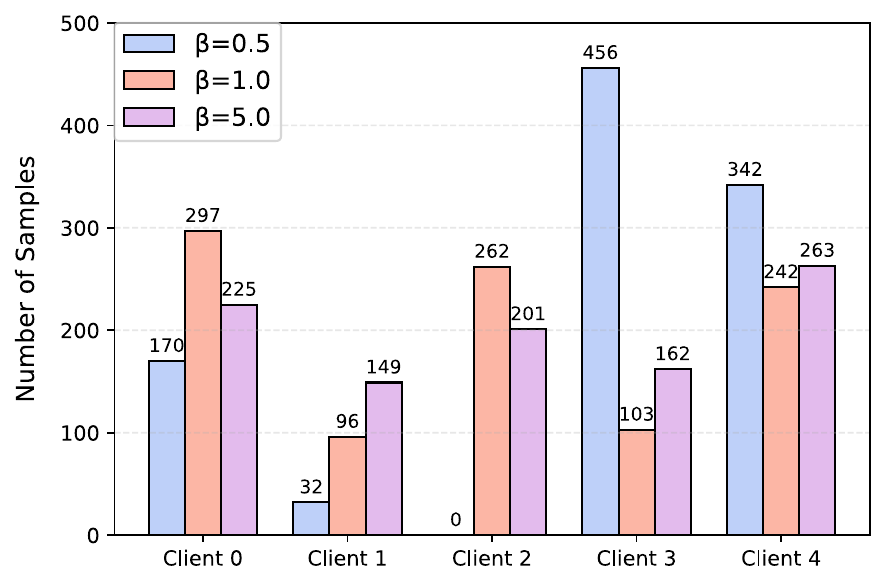}
    \vspace{-5pt}
    \caption{Visualization of Dirichlet distribution.}
    \label{fig:beta}
\vspace{-10pt}
\end{figure}

In federated learning tasks, data heterogeneity poses a core challenge in distributed training, primarily manifesting as varying proportions of private data across different clients. Therefore, we employ the Dirichlet distribution, a common approach in FL tasks, to model distributional differences among clients. In this paper, we use $\beta$ to control the degree of distributional variation, as visualized in \Cref{fig:beta}.

It can be seen that a smaller $\beta$ leads to greater disparities in data distribution among clients, resulting in more extreme heterogeneity, whereas a larger value of $\beta$ indicates a more uniform distribution.

\subsection{Visualization of the dataset}
\Cref{tab:visualization1} and~\Cref{tab:visualization2} present the input images and instruction formats of the 12 selected datasets, which exhibit relatively low average zero-shot performance on the base model LLaVA-v1.5-7B, approximately \textbf{30\%} lower than their fine-tuning performance~(See Zero-shot and Centralized MTL in Table~\textcolor{iccvblue}{4}). This better ensures that these datasets remain unseen or unfamiliar to the base model during training, thereby reducing information leakage.

In our experiments, we design two dataset-level settings: Capability-related and Task-related. The capability-related setting categorizes the 12 datasets into four dimensions: general, math, chart, and other, where each capability consists of a mix of relevant datasets. The task-related setting selects 8 out of the 12 datasets for different stages of continual learning, evaluating how well different approaches mitigate forgetting over a long-stage learning setup. The specific data composition of these two settings can be found in Section~\textcolor{iccvblue}{3.2}.

\begin{table*}
    \begin{minipage}{0.99\textwidth}
    \setlength{\tabcolsep}{4pt}
    \centering
    \resizebox{\linewidth}{!}{
    \begin{tabular}{c|ccc}
    \toprule
    \textbf{Dataset} & \textbf{Visual input} & \textbf{Question} & \textbf{Response} \\ 
    \midrule
    A-OKVQA & \adjustbox{valign=c}{\includegraphics[height=2.5cm, width=2.5cm]{figure/source/AOKVQA.pdf}} & \makecell{What is the man by the bags awaiting?\\0. skateboarder 1. train 2. delivery 3. cab \\ Answer with the option's letter from the given choices directly.} & 3 \\
    \midrule
    IconQA & \adjustbox{valign=c}{\includegraphics[height=2.5cm, width=3.5cm]{figure/source/iconqa.pdf}} & \makecell{What has been done to this letter? \\ 0. turn 1. slide 2. flip \\ Answer with the option's letter from the given choices directly.} & 2 \\
    \midrule
    Grounding & \adjustbox{valign=c}{\includegraphics[height=2.5cm, width=1.5cm]{figure/source/grounding.pdf}} & \makecell{Please provide the bounding box coordinate \\of the region this sentence describes: \\ kid on right teddy bear bib.}  & [0.65,0.44,0.88,0.98] \\
    \midrule
    ImageNet-R & \adjustbox{valign=c}{\includegraphics[height=2.5cm, width=2.5cm]{figure/source/imagenet_r.pdf}} &\makecell{ Question: What is the object in the image? \\ Answer the question using a single word or phrase.} & Killer whale \\
    \midrule
     ArxivQA  &  \adjustbox{valign=c}{\includegraphics[height=3.0cm, width=3.5cm]{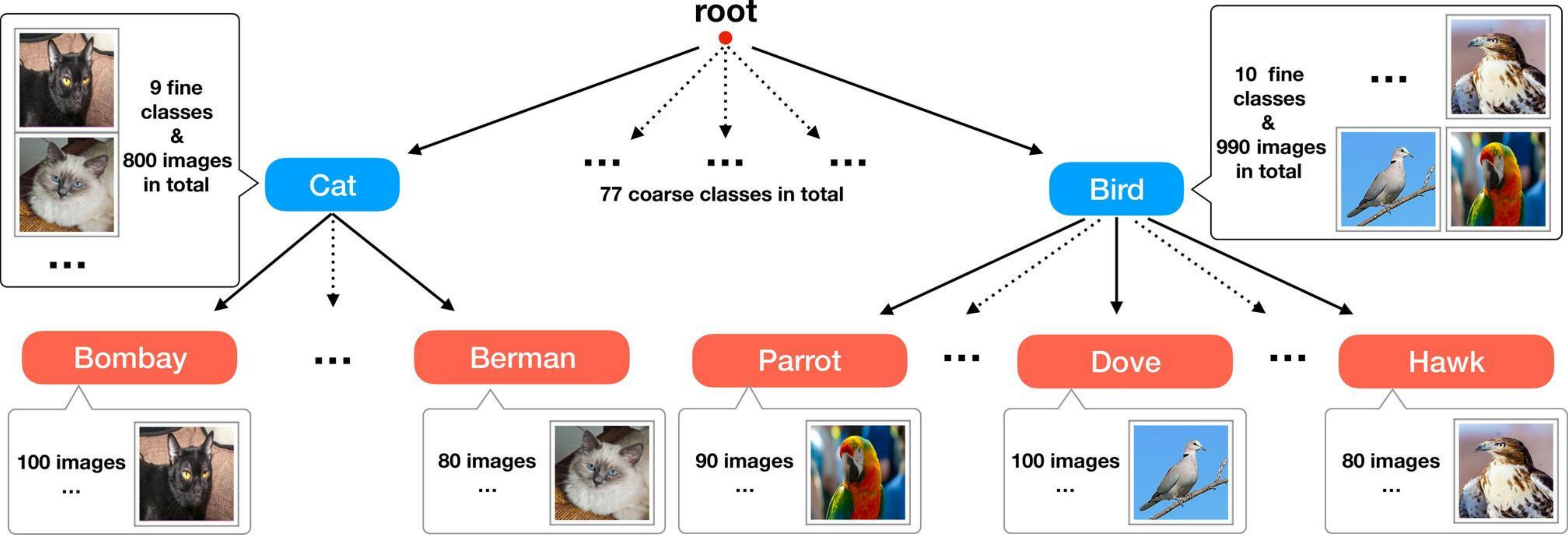}}  & \makecell{How many coarse classes are represented in the figure? \\ A) Less than 50 ~B) Exactly 77 \\ C) More than 100 D) Exactly 99  \\  Answer with the option's letter from the given choices directly. } &  B  \\
     \midrule
    FigureQA & \adjustbox{valign=c}{\includegraphics[height=3.0cm, width=3.0cm]{figure/source/figureqa.pdf}} & \makecell{Please answer the question and provide the correct option letter at the end. \\ Question: Is Light Slate greater than Dark Green? \\ Choices: (A) no (B) yes}  & B \\
    \midrule
    DVQA & \adjustbox{valign=c}{\includegraphics[height=2.5cm, width=2.5cm]{figure/source/dvqa.pdf}} & \makecell{Please answer the question and provide the final answer at the end.\\ Question: What is the value of the largest individual bar in the whole chart?} & 9 \\
    \midrule
    CLEVR-Math & \adjustbox{valign=c}{\includegraphics[height=2.5cm, width=2.5cm]{figure/source/clevr.pdf}} & \makecell{Subtract all brown matte objects. \\Subtract all blue cylinders. How many objects are left?\\ Answer the question using a single word or phrase. } & 7 \\
    \midrule
    super-CLEVR & \adjustbox{valign=c}{\includegraphics[height=2.5cm, width=2.5cm]{figure/source/super.pdf}} & \makecell{Question: There is a matte thing that is in front of the small \\ purple utility bike and behind the red metal thing; how big is it?\\ Answer the question using a single word or phrase.} & small \\
    \bottomrule
    \end{tabular}
    }
    \caption{Visualization of input images and instruction formats for each dataset in FCIT.}
    \label{tab:visualization1}  
    \end{minipage}
\end{table*}

\begin{table*}[t]
    \begin{minipage}{0.99\textwidth}
    \setlength{\tabcolsep}{4pt}
    \centering
    \resizebox{\linewidth}{!}{
    \begin{tabular}{c|ccc}
    \toprule
    \textbf{Dataset} & \textbf{Visual input} & \textbf{Question} & \textbf{Response} \\ 
    \midrule
    TabMWP & \adjustbox{valign=c}{\includegraphics[height=3.0cm, width=3.0cm]{figure/source/tabmwp.pdf}} & \makecell{Question: The members of the local garden club \\ tallied the number of plants in each person's garden. \\ How many gardens have at least 47 plants?\\ Answer the question using a single word or phrase.} & 13 \\
    \midrule
    Flickr30k & \adjustbox{valign=c}{\includegraphics[height=3.0cm, width=3.0cm]{figure/source/flickr30k.pdf}} & \makecell{What is happening in the image?\\ Generate a brief caption for the image.}  & \makecell{Three older women are at a restaurant \\ talking with other people} \\
    \midrule
    OCR-VQA & \adjustbox{valign=c}{\includegraphics[height=3.5cm, width=2.0cm]{figure/source/ocrvqa.pdf}} & \makecell{Who wrote this book? \\ Answer the question using a single word or phrase.} & Tony Wheeler \\
    \bottomrule
    \end{tabular}
    }
    \caption{Visualization of input images and instruction formats for each dataset in FCIT.}
    \label{tab:visualization2}  
    \end{minipage}
\end{table*}

\setcounter{figure}{0}
\renewcommand{\thefigure}{B\arabic{figure}}
\renewcommand{\thetable}{B\arabic{table}}
\section{More Details of Experiments}
\label{appendix:B}

\subsection{Details of the comparison method}
In this section, we present the underlying principles of the baseline methods used in our experiments.

\textbf{LwF} mitigates forgetting by applying knowledge distillation loss during new task learning. It preserves past knowledge by extracting soft labels from the frozen old model’s outputs and constraining the new model’s outputs to remain close, minimizing deviation from previous tasks.

\textbf{EWC} mitigates forgetting by restricting updates to important parameters of previous tasks. It computes parameter importance using the Fisher information matrix and penalizes significant changes, preserving past knowledge while learning new tasks.

\textbf{L2P} introduces a dynamic prompts pool, enabling the model to select and optimize relevant prompts based on similarity during training. Additionally, it applies a regularization loss to encourage task-specific prompt selection, mitigating catastrophic forgetting.

\textbf{O-LoRA} imposes an orthogonality constraint in parameter space, ensuring that the optimization of the current task occurs in a direction orthogonal to previous tasks, thereby minimizing task conflicts. During inference, it aggregates learned knowledge by concatenating the LoRA modules of all tasks along the specified dimension.

\textbf{M-LoRA} trains LoRA modules separately at each stage and mitigates forgetting by spatially merging them in parameter space during inference. Unlike O-LoRA, it does not incur additional memory overhead during training.

\textbf{MoELoRA} transforms the fine-tuning of individual LoRA into a Mixture-of-Experts framework, where a predefined set of LoRA modules serves as expert heads. During training, routers are optimized alongside expert selection, aiming to assign different tasks to distinct expert heads. This structured allocation helps mitigate forgetting by ensuring the router effectively distributes outputs across expert modules.

\subsection{Details of evaluation}
In our benchmark, tasks have different output formats, requiring tailored accuracy evaluation methods. For tasks with answering a single option or single word, we determine correctness using \texttt{pred.upper() in Response.upper()}. For captioning tasks, we adopt standard image captioning metrics, including \texttt{Bleu\_1}, \texttt{Bleu\_2}, \texttt{Bleu\_3}, \texttt{Bleu\_4}, \texttt{METEOR}, \texttt{ROUGE\_L}, and \texttt{CIDEr}. The final results are computed as the average of these seven metrics.

\subsection{Ablation study of hyper-parameters}
\begin{figure}[t]
    \centering
    \includegraphics[width=0.7\linewidth]{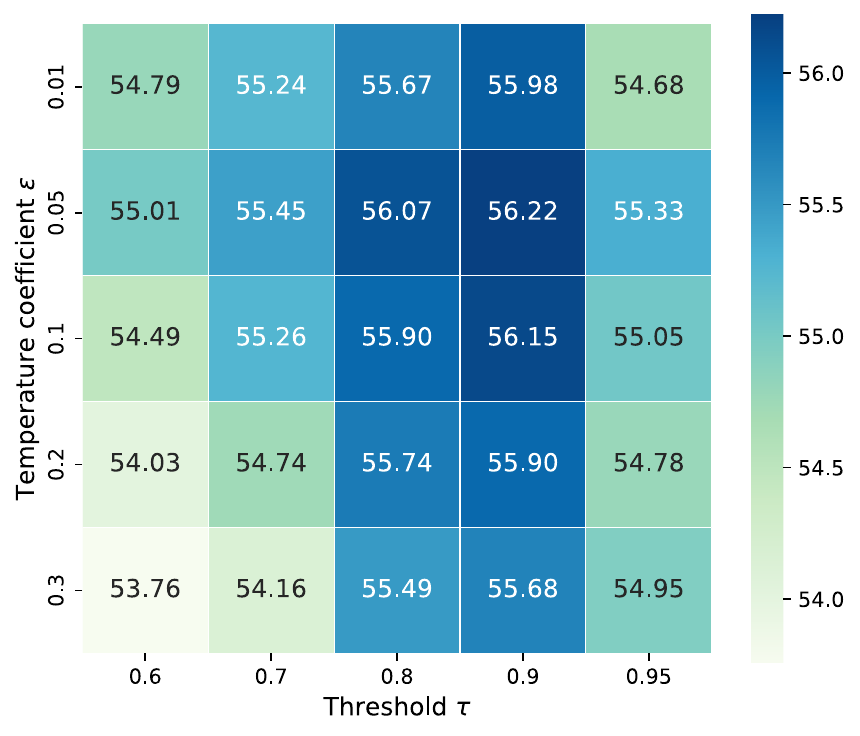}
    \vspace{-5pt}
    \caption{Ablation study of hyper-parameters in Hom-FCIT and task-related setting. The partition $\beta$ is set to 1.0.}
    \label{fig:hetamap}
\vspace{-10pt}
\end{figure}

In this section, we conduct more ablation experiments on two key hyperparameters, the threshold $\tau$ and the temperature coefficient $\varepsilon$, with results shown in \Cref{fig:hetamap}. It can be seen that both excessively large and small temperature coefficients significantly affect the results. A larger coefficient leads to overly sharp activation assignments, increasing the likelihood of selecting mismatched subspaces, while a smaller coefficient incorporates excessive information from irrelevant subspaces, ultimately degrading model performance. Similarly, an excessively large threshold may filter out knowledge relevant to the same task, while a too-small threshold may misassign identity tokens to other subspaces, both leading to degraded performance.

\begin{table}[h]
    \centering
    \resizebox{\linewidth}{!}{
    \begin{tabular}{c|ccccccc}
    \toprule
        Threshold $\tau$ & $\leq$~0.5 & 0.6 & 0.7 & 0.75 & 0.8 & 0.9 & 0.95\\
       \midrule
        Num. of Subspace  & 1 & 3 & 6 & 7 & 8 & 8 & 10\\
    \bottomrule
    \end{tabular}}
    \caption{The effect of threshold selection on the number of subspaces in Hom-FCIT setting (task-related, $\beta=1.0$).}
    \vspace{-10pt}
    \label{tab:subspaces}
\end{table}

We further investigate the effect of threshold selection on the number of subspaces formed. As shown in Table~\ref{tab:subspaces}, a threshold that is too low fails to effectively separate task-specific knowledge, leading to subspace aggregation across tasks and potential knowledge conflicts. Conversely, an overly high threshold may split knowledge that belongs to the same subspace, resulting in redundant branches and degraded model performance.

\subsection{Comparison on a single dataset}
\begin{figure}[h]
    \centering
    \includegraphics[width=0.95\linewidth]{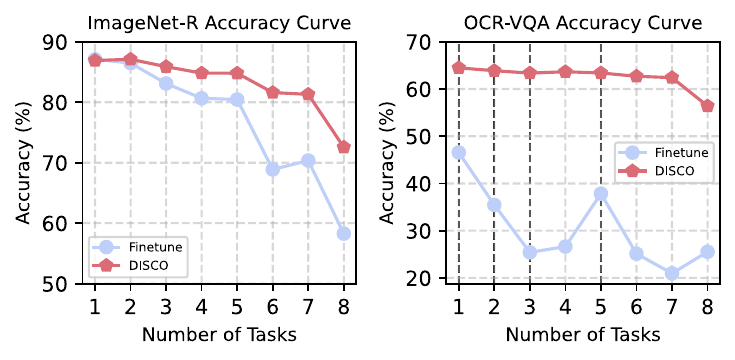}
    \vspace{-5pt}
    \caption{\textbf{Left}. Performance curve of first learned task ImageNet-R under Hom-FCIT and task-related settings; \textbf{Right}. Performance curve of OCR-VQA under Het-FCIT and task-related settings. The black dashed line indicates the stage where the model has learned OCR-VQA.}
    \label{fig:single}
\vspace{-10pt}
\end{figure}

We further compare single-task performance between our proposed DISCO and Finetune. As shown in \Cref{fig:single} (Left), for the first learned task, our method significantly mitigates forgetting. Under Het-FCIT (\Cref{fig:single} Right), for task-specific OCR-VQA, DISCO not only enhances knowledge retention during learning but also maintains strong performance even when the task is absent. In contrast, Finetune suffers from severe inter-task conflicts, leading to continuous performance degradation even while learning the dataset (2-nd and 3-rd black dashed lines).

\subsection{Evaluation of zero-shot capability and general benchmark performance}

In addition to mitigating forgetting, we also evaluate the generalization ability of different methods on unseen tasks, as well as their performance on a general LMM benchmark. As shown in Table~\ref{tab:tableB2}, our method improves zero-shot transfer performance on tasks without downstream supervision. Moreover, it minimizes negative transfer effects on the general LMM benchmark, demonstrating strong robustness and adaptability.

\begin{table}[h]
    \centering
    \resizebox{0.95\linewidth}{!}{
    \begin{tabular}{c|cccc|cccc}
    \toprule
        \multirow{2}{*}{\makecell{\textbf{Hom-FCIT} \\ $\beta=1.0$}} & \multicolumn{4}{c||}{\textbf{Zero-shot capability}} & \multicolumn{4}{c}{\textbf{General LMM benchmark}} \\
        \cmidrule{2-9}
        & Task2 & Task3 & \multicolumn{1}{c}{Task4} & \multicolumn{1}{c||}{Avg} & MME & POPE & MMBench & SEED \\
        \midrule
        LLaVA-1.5 & \textbf{46.97} & \textbf{17.05} & \multicolumn{1}{c}{27.13} & \multicolumn{1}{c||}{30.38} & \textbf{1476.9} & \textbf{86.4} & \textbf{66.1} & \underline{60.1} \\
        O-LoRA & 41.89 & 16.04 & \multicolumn{1}{c}{26.88} & \multicolumn{1}{c||}{28.27} & 1354.6 & 80.7 & 59.4 & 59.0 \\
        M-LoRA & 45.33 & 15.90 & \multicolumn{1}{c}{\underline{31.26}} & \multicolumn{1}{c||}{\underline{30.83}} & 1376.8 & 79.8 & 60.1 & 59.5 \\
        MoELoRA & 43.16 & 16.86 & \multicolumn{1}{c}{29.09} & \multicolumn{1}{c||}{29.70} & 1358.3 & 82.8 & 60.5 & 59.2  \\
        \textbf{DISCO} & \underline{46.14} & \underline{17.00} & \multicolumn{1}{c}{\textbf{36.09}} & \multicolumn{1}{c||}{\textbf{33.08}} & \underline{1436.6} & \underline{83.9} & \underline{62.4} & \underline{60.1}  \\
    \bottomrule
    \end{tabular}}
    \caption{Zero-shot transfer and general LMM benchmark results in the \textbf{Hom-FCIT} setting (capability-related, $\beta=1.0$).}
    \label{tab:tableB2}
\vspace{-10pt}
\end{table}

\subsection{Efficiency and storage analysis.}

\begin{table}[h]
    \centering
    \resizebox{0.95\linewidth}{!}{
    \begin{tabular}{c|cc|cccc}
    \toprule
        \multirow{2}{*}{Methods} & \multicolumn{2}{c||}{\textbf{Efficiency}} & \multicolumn{4}{c}{\textbf{LoRA memory cost}} \\
        \cmidrule{2-7}
        & Speed~($\uparrow$) & \multicolumn{1}{c||}{FLOPs~($\downarrow$)} & \multicolumn{1}{c}{Hom-FCIT} & \multicolumn{1}{c}{$\Delta$} & Het-FCIT & $\Delta$  \\
        \midrule
        Finetune & \textbf{3.46~it/s} & \multicolumn{1}{c||}{\textbf{8.55~T}} & \textbf{22.2~M} & \textbf{0.16\%} & \textbf{22.2~M} & \textbf{0.16\%} \\
        O-LoRA & 3.42~it/s & \multicolumn{1}{c||}{\underline{9.35~T}} & \underline{88.7~M} & \underline{0.63\%} & \underline{177.0~M} & \underline{1.26\%} \\
        MoELoRA & 3.36~it/s & \multicolumn{1}{c||}{9.77~T} & 93.4~M & 0.68\% & 186.0~M & 1.33\% \\
        \textbf{DISCO} & \underline{3.45~it/s} & \multicolumn{1}{c||}{\underline{9.35~T}} & 88.9~M & \underline{0.63\%} & 178.0~M & \underline{1.26\%} \\
    \bottomrule
    \end{tabular}}
    \caption{All efficiency comparisons are conducted under identical conditions. $\Delta$: relative percentage to the backbone model.}
    \label{tab:tableB3}
\vspace{-10pt}
\end{table}

As shown in Table~\ref{tab:tableB3}, our method demonstrates the most substantial performance gains compared to fine-tuning and other baselines, while preserving competitive inference efficiency and requiring only modest additional LoRA storage overhead.

\subsection{Detailed Results of DISCO.}

\begin{table*}[h]
    \centering
    \resizebox{0.9\linewidth}{!}{
    \begin{minipage}{0.4\linewidth}
    \centering
    \resizebox{\linewidth}{!}{
    \begin{tabular}{l|cccc}
    \toprule
       DISCO & General & Other & Chart & Math \\
       \midrule
       Task1  & 71.26 & & & \\
       Task2  & 67.72 & 57.23 & & \\
       Task3  & 64.47 & 52.75 & 50.10 & \\
       Task4  & 62.92 & 53.14 & 46.43 & 56.15 \\
    \bottomrule
    \end{tabular}}
    \caption{Results matrix of DISCO in Hom-FCIT setting (capability-related, $\beta=1.0$)}
    \label{tab:tableB4}
    \end{minipage}
    \qquad
    \begin{minipage}{0.4\linewidth}
    \centering
    \resizebox{\linewidth}{!}{
    \begin{tabular}{l|cccc}
    \toprule
       DISCO & General & Other & Chart & Math \\
       \midrule
       Task1  & 69.87 & 55.51 & & \\
       Task2  & 66.46 & 54.75 & 47.81 & 51.87 \\
       Task3  & 68.73 & 56.30 & 50.00 & 54.47 \\
       Task4  & 68.01 & 55.49 & 54.19 & 57.88 \\
    \bottomrule
    \end{tabular}}
    \caption{Results matrix of DISCO in Het-FCIT setting (capability-related, $\beta=1.0$)}
    \label{tab:tableB5}
    \end{minipage}
    }
\end{table*}

\begin{table*}[t]
    \centering
    \resizebox{0.9\linewidth}{!}{
    \begin{tabular}{l|cccccccc}
    \toprule
       DISCO & ImageNet-R & ArxivQA & IconQA & CLEVR-Math & OCRVQA & Flickr30k & FigureQA & super-CLEVR \\
       \midrule
       Task1  & 86.90 &  &  &  &  &  &  &  \\
       Task2  & 87.10 & 93.40 &  &  &  &  &  &  \\
       Task3  & 85.88 & 93.35 & 65.47 &  &  &  &  &  \\
       Task4  & 84.81 & 93.79 & 58.70 & 60.28 &  &  &  &  \\
       Task5  & 84.80 & 93.96 & 59.14 & 60.84 & 63.76 &  &  &  \\
       Task6  & 81.60 & 93.02 & 58.62 & 56.24 & 37.22 & 54.57 &  &  \\
       Task7  & 81.30 & 92.45 & 58.27 & 56.03 & 25.82 & 54.46 & 43.70 &  \\
       Task8  & 72.56 & 92.34 & 55.59 & 47.72 & 35.97 & 52.49 & 42.92 & 50.16 \\
    \bottomrule
    \end{tabular}}
    \caption{Results matrix of DISCO in Hom-FCIT setting (task-related, $\beta=1.0$)}
    \label{tab:tableB6}
\end{table*}

\begin{table*}[t]
    \centering
    \resizebox{0.9\linewidth}{!}{
    \begin{tabular}{l|cccccccc}
    \toprule
       DISCO & ImageNet-R & ArxivQA & IconQA & CLEVR-Math & OCRVQA & Flickr30k & FigureQA & super-CLEVR \\
       \midrule
       Task1  &  & 87.99 & 58.37 &  & 64.76 & 53.39 & 41.52 &  \\
       Task2  &  & 88.35 & 54.87 & 53.68 & 64.35 & 53.52 & 42.08 & 47.04 \\
       Task3  & 76.46 & 88.1 & 61.35 & 57.22 & 64.24 & 55.98 & 41.85 & 46.88 \\
       Task4  & 84.45 & 89.28 & 61.27 & 55.33 & 64.04 & 55.94 & 40.75 & 50.00 \\
       Task5  & 83.90 & 90.45 & 63.83 & 58.23 & 64.07 & 53.97 & 39.92 & 50.18 \\
       Task6  & 85.46 & 90.71 & 64.14 & 62.53 & 63.73 & 53.68 & 40.23 & 50.14 \\
       Task7  & 84.75 & 90.65 & 68.36 & 60.28 & 63.17 & 42.90 & 40.38 & 51.08 \\
       Task8  & 73.95 & 91.75 & 68.33 & 59.55 & 62.52 & 55.18 & 43.55 & 51.16 \\
    \bottomrule
    \end{tabular}}
    \caption{Results matrix of DISCO in Het-FCIT setting (task-related, $\beta=1.0$)}
    \label{tab:tableB7}
\end{table*}

\end{document}